\documentclass{article}

\usepackage[preprint]{styleFile}
\usepackage{algorithm}
\usepackage{algorithmic}

\usepackage[utf8]{inputenc} 
\usepackage[T1]{fontenc}    
\usepackage{hyperref}       
\usepackage{url}            
\usepackage{booktabs}       
\usepackage{amsfonts}       
\usepackage{nicefrac}       
\usepackage{microtype}      
\usepackage{xcolor}         
\usepackage[pdftex]{graphicx}
\usepackage{amsmath}

\title{Explanation of Dynamic Physical Field Predictions using WassersteinGrad: Application to Autoregressive Weather Forecasting}

\author{
Younes Essafouri\textsuperscript{1} \\
\texttt{younes.essafouri@cnrs.fr} \\
\And
Laure Raynaud\textsuperscript{2} \\
\texttt{laure.raynaud@meteo.fr} \\
\And
Luciano Drozda\textsuperscript{3} \\
\texttt{drozda@cerfacs.fr} \\
\And
Laurent Risser\textsuperscript{1} \\
\texttt{laurent.risser@cnrs.fr} \\
\\
\textsuperscript{1} Univ. Toulouse, INSA Toulouse, CNRS UMR 5219, IMT, Toulouse, France \\
\textsuperscript{2} Météo-France, CNRS, Univ. Toulouse, CNRM, Toulouse, France \\
\textsuperscript{3} Cerfacs, CNRS/Cerfacs/IRD, CECI, Toulouse, France
}

\begin{document}

\maketitle

\begin{abstract}
As the demand to integrate Artificial Intelligence into high-stakes environments continues to grow, explaining the reasoning behind neural-network predictions has shifted from a theoretical curiosity to a strict operational requirement. Our work is motivated by the explanations of autoregressive neural predictions on dynamic physical fields, as in weather forecasting. Gradient-based feature attribution methods are widely used to explain the predictions on such data, in particular due to their scalability to high-dimensional inputs. 
It is also interesting to remark that gradient-based techniques such as SmoothGrad are now standard on images to robustify the explanations using pointwise averages of  the attribution maps obtained from several noised inputs. 
Our goal is to efficiently adapt this aggregation strategy to dynamic physical fields. To do so, our first contribution is to identify a fundamental failure mode when averaging perturbed attribution maps on dynamic physical fields: stochastic input perturbations do not induce stationary amplitude noise in attribution maps, but instead cause a \emph{geometric displacement} of the attributions. Consequently, pointwise averaging blurs these spatially misaligned features. 
To tackle this issue, we  introduce \emph{WassersteinGrad}, which extracts a geometric consensus of perturbed attribution maps by computing their entropic Wasserstein barycenter.
The results, obtained on regional weather data and a meteorologist-validated neural model,  demonstrate promising explainability properties of WassersteinGrad over gradient-based baselines across both single-step and autoregressive forecasting settings.
\end{abstract}

\section{Introduction}
Deep learning models have achieved remarkable complexity and performance across a wide array of tasks. However, their inherent opaque nature often remains a critical barrier to operational trust in various domains~\citep{623216, lipton2018mythos}. This opacity is especially problematic when deploying deep neural models in high-stakes applications, where erroneous predictions can have severe physical or societal consequences. A striking example that motivated our work is weather forecasting, where deep learning is increasingly competing with traditional physical solvers~\citep{moldovan2025update,lam2023learning,alet2025skillful,bommer2024finding}. As these systems approach deployment in high-stakes decision pipelines, a fundamental question that arises to trust their predictions is: \textit{Which atmospheric features drive given forecasts?}  Explainable AI (XAI) aims to address this transparency gap by systematically revealing the decision-making processes behind model predictions~\citep{arrieta2020explainable}.

Among existing XAI strategies, gradient-based feature attribution methods have become standard for highlighting salient input regions related to a prediction, when the inputs are sampled on a regular and high dimensional spatial domain~\citep{wang2024gradient,mehrpanah2024spectral}. However, a fundamental limitation of utilizing raw input gradients to explain deep-learning-based predictions is that they are often visually noisy and locally shattered due to the highly non-linear nature of deep networks~\citep{balduzzi2017shattered}. To reduce this noise, widely adopted smoothing techniques like SmoothGrad \citep{smilkov2017smoothgrad} first sample several noisy neighbors of the input observation with a spatially independent Gaussian noise, and then average the gradients explaining the predictions obtained using the perturbed inputs.
Although these strategies have proven to be particularly effective on image data,  we argue that the reliance on pointwise averaging leads to poorly localized explanations, when used to forecast the state of dynamic physical phenomena.

In our driving application, meteorological phenomena are characterized by their continuous structural geometry and advective movement, as formalized by the governing fluid dynamics (see \textit{e.g.}~\citep{kalnay2003atmospheric}, pages 56--60). 
When explaining the predictions of noisy meteorological inputs, it appeared to us that the attribution masses tend to migrate from their true spatial location to nearby and physically incorrect locations. This is formally recognized as phase error in meteorological verification~\citep{DistortionRepresentationofForecastErrors,ebert2008, keil2007}, predicting the right thing but in the wrong location. We showcase an empirical evidence for this phenomenon in Figure~\ref{fig:intro}(c).

This failure is not incidental. We formally demonstrate in our paper that this is compounded by the intrinsic structure of the layers that are commonly used in deep neural forecasting models. In addition,  we  empirically show that the geometric displacement of the explanations is reinforced when the prediction model is used in an autoregressive manner, as illustrated in Figure~\ref{fig:intro}(a), which has become standard in weather forecasting~\citep{lam2023learning,moldovan2025update}. Each autoregressive step likely introduces an independent geometric distortion. When backpropagating gradient from final lead times to explain noisy inputs, these spatial misalignments accumulate. 
This makes it particularly interesting to develop novel XAI solutions that take into account the intrinsic properties  of modern autoregressive forecast models. This is what we do in our paper, with the introduction of the transport-based aggregation framework \emph{WassersteinGrad}.

\begin{figure}[h]
  \centering
  \includegraphics[width=1.0\linewidth]{figures/introduction/introduction_figure_paper3.pdf}
 \caption{
\textbf{(a)} Autoregressive use of a neural-based forecasting model $f$, producing a sequence of physical fields (here regional atmospheric states with $C=21$ channels) at successive lead times $t+1, \ldots, t+T$. 
\textbf{(b - from left to right)}  
input zonal wind at 250\,hPa at time $t$; predicted surface precipitation at time $t+5$; explanation of rain prediction in Paris area using the  gradient attribution method of \citep{baehrens2010explain}.
\textbf{(c~-~top)}~Gradients obtained on baseline and perturbed input data for $T=1$.
\textbf{(c~-~bottom)}~impact of the noise on the spatial motion of gradient information (see Section~\ref{sec:method}). Note that although the injected noise has a perciptible effect on the gradient-based attribution map, it only induced about 1$\%$ of relative prediction error.
}
\label{fig:intro}
\end{figure}

Our main contributions are as follows:
\begin{enumerate}
    \item \textbf{Mechanistic analysis of gradient displacement:} We provide an in-depth discussion of why input perturbations induce geometric rather than stationary noise in attribution maps. By tracing noise propagation through two ubiquitous architectural components: \textbf{(i)} pooling argmax instability in CNNs, which physically relocates backpropagated gradients, and \textbf{(ii)} attention switching in Transformers, formalized via mean-field interacting particle dynamics~\citep{geshkovski2025mathematical,rigollet2025mean}.  We identify structural inductive biases that make geometric displacement of backpropagated gradients architecturally expected.

    \item \textbf{Transport based Aggregation Framework:} We propose \textbf{WassersteinGrad}, a transport-based aggregation framework.
    By mapping raw gradients into the space of spatial probability measures, we utilize the entropic Wasserstein barycenter~\citep{doi:10.1137/100805741, pmlr-v32-cuturi14} to extract a \textit{geometric consensus} of perturbed explanations.
    \item \textbf{Empirical Validation on Neural Weather Forecasting:} We provide the first empirical characterization of geometric displacement in gradient attributions for standard and autoregressive neural weather forecasting models. We first use an optimal transport mass flux analysis to show that the deformation of attribution maps is coherent and spatially structured. We then show that this manifests in practice as a spatial displacement of the attribution centroid, with high stochastic variance. Finally, we compare \textit{WassersteinGrad} with widely used gradient-based baselines across faithfulness, robustness, and sparsity metrics, demonstrating consistent improvements.
             
\end{enumerate}
\section{Related Works}

\subsection{Gradient-Based Methods and Smoothing}
The gradient of a model's prediction reflects its sensitivity to infinitesimal changes in the input space. Methods such as Saliency~\citep{baehrens2010explain,simonyan2013deep} and Input$\times$Gradient~\citep{ancona2017towards,shrikumar2016not} assign importance scores to input features by computing the partial derivatives of the output with respect to the input. However, the resulting gradient maps are often visually noisy and locally discontinuous~\citep{hill2025smoothed,smilkov2017smoothgrad}. As highlighted by~\citep{balduzzi2017shattered}, this noise emerges due to the high local variation of gradients in deep networks with non-linear activations (e.g., ReLU), giving rise to the Shattered Gradients Problem (SGP)~\citep{balduzzi2017shattered}. 
Finally, Integrated Gradients~\citep{sundararajan2017axiomatic} computes attributions by integrating gradients along weighted averages between the input and a baseline input, assumed to be neutral. It, however, requires defining such a baseline, and the resulting intermediate inputs may be physically meaningless in non-linear physical settings. 

\paragraph{SmoothGrad.}

To mitigate the SGP,~\citep{smilkov2017smoothgrad} 
approximate a Gaussian-smoothed gradient via Monte Carlo sampling over $N$ noisy 
inputs $\tilde{x}_i \sim \mathcal{N}(x, \sigma^2)$:
\[
G_{\text{SG}}(x) = \frac{1}{N} \sum_{i=1}^{N} \nabla f(\tilde{x}_i)
\]
Other popular methods extend SmoothGrad along 
different axes: NoiseGrad and 
FusionGrad~\citep{bykov2022noisegrad} introduce 
stochasticity into model weights rather than 
inputs, and VarGrad~\citep{adebayo2018local} 
replaces the mean with the variance of perturbed 
gradients. None of these methods, however, specifically address the geometric displacement we identify.

\subsection{The Fragility of Explanations and Spatial Uncertainty}

Recent literature has exposed fundamental limitations in gradient-based smoothing. For instance, some argue that SmoothGrad acts as a band-pass filter, suppressing both high and low-frequency components~\citep{mehrpanah2024spectral}. This can lead to spectral bias and inconsistencies, where the resulting explanations change drastically with hyperparameter choices (e.g., the noise variance $\sigma^2$). Due to noise sampling,~\citep{zhou2025adaptgrad} shows that SmoothGrad suffers from out of distribution error. While the empirical fragility of attribution maps is well-documented, the theoretical mechanics by which input noise propagates through specific architectures remain under-explored in XAI. Existing theoretical works primarily analyze forward-pass noise propagation in the context of adversarial robustness~\citep{pmlr-v97-cohen19c} or study gradient magnitudes (\textit{e.g.}, exploding/vanishing gradients)~\citep{pascanu2013difficulty}. Critically, all these works conceptualize gradient noise as a \emph{stationary} phenomenon: noise affects attribution \emph{amplitude} at each spatial location independently. On the architectural side,~\citep{kim2021lipschitz} demonstrated that standard attention layers lack Lipschitz continuity, making their gradients inherently unstable. To formally understand these dynamics, recent studies have elegantly modeled Transformer attention mechanisms~\citep{vaswani2017attention} as interacting particle systems~\citep{rigollet2025mean,geshkovski2025mathematical,bruno2025a}. In this work, we build upon this interacting-particle framework to analyze how architectural inductive biases transform additive input noise into geometric spatial deformation in the attention layer.

\subsection{Optimal Transport and Wasserstein Barycenters}
The Wasserstein metric measures the minimal transport cost between probability measures, incorporating the underlying geometry of the 
space~\citep{alma991005863149705596}. Within XAI, OT has primarily been used  as a diagnostic tool: \citep{Rhodes_2024_CVPR} leverage it to compare saliency maps, and \citep{naumann2026wasserstein} use it to identify features driving dataset shifts. To \emph{aggregate} distributions, \citep{doi:10.1137/100805741} introduced the Wasserstein barycenter which generalizes the concept of an average by finding the Fréchet mean in Wasserstein space. Unlike pointwise averaging, the Wasserstein barycenter preserves geometric structure~\citep{pmlr-v32-cuturi14}. Made tractable via entropic regularisation and Sinkhorn iterations~\citep{pmlr-v32-cuturi14}, Wasserstein barycenter has been applied in graphics~\citep{10.1145/2766963},clustering~\citep{ye2017fast}, and model ensembling~\citep{dognin2019wasserstein}.

\section{Methodology}
\subsection{Pipeline for Explainability and Attribution Setup}
\label{sec:pipeline}

Let $x_t \in \mathbb{R}^{H \times W \times C}$ denote the physical field 
state at time $t$, defined on a spatial grid of $H \times W$ points 
with $C$ input channels (\textit{e.g.}\ wind speed, temperature, 
pressure). Let $f: \mathbb{R}^{H \times W \times C} \to 
\mathbb{R}^{H \times W \times C}$ denote the neural forecasting model, 
applied autoregressively over $T$ steps:
\begin{equation}
    \hat{x}_{t+T} = f^T(x_t) = 
    \underbrace{f \circ f \circ \cdots \circ f}_{T}(x_t).
\end{equation}
As illustrated in Figure \ref{fig:intro}(b), we use 
the following four-step pipeline to generate localised attributions:

\begin{enumerate}
    \item \textbf{Input channel selection.} We select a single input 
    channel $c_{\mathrm{in}} \in \{1, \ldots, C\}$ on which spatial 
    attributions will be represented (\textit{e.g.}\ zonal wind at 
    250 hPa). We denote the corresponding spatial slice as 
    $x_{\mathrm{in}} = [x_t]_{c_{\mathrm{in}}} \in 
    \mathbb{R}^{H \times W}$.

    \item \textbf{Output channel selection.} We select a single 
    output channel $c_{\mathrm{out}} \in \{1,\ldots,C\}$ of the predicted field to 
    explain (\textit{e.g.}\ 10~m surface wind speed):
    $\mathcal{Y}_{\mathrm{out}} = 
        \left[f^T(x_t)\right]_{c_{\mathrm{out}}} \in \mathbb{R}^{H \times W}.$
    Gradients backpropagate through all $T$ compositions of $f$.

   \item \textbf{Region of Interest (ROI).} We define a spatial 
    bounding box $\mathcal{B}$ over a region of interest 
    (\textit{e.g.}\ Paris). We then define an 
    aggregation operator $\phi: \mathbb{R}^{H \times W} \to 
    \mathbb{R}$ that maps the forecast field to a scalar target. 
    In this work, $\phi$ is the spatial average over $\mathcal{B}$:
$    \mathcal{Y}_{\mathrm{target}} = \phi\!\left(
        \mathcal{Y}_{\mathrm{out}}\right) = \frac{1}{|\mathcal{B}|} 
        \sum_{u \in \mathcal{B}} \mathcal{Y}_{\mathrm{out}}(u).$
        Note that $\phi$ can be any differentiable aggregation operator. 

    \item \textbf{Attribution computation.} We compute the 
    attribution map as the gradient of the scalar target with 
    respect to the selected input channel:
        $G = \nabla_{x_{\mathrm{in}}}\, \mathcal{Y}_{\mathrm{target}} 
        \in \mathbb{R}^{H \times W}.$
        
\end{enumerate}

To simplify notation in the following sections, we drop the time index $t$ and denote the input as $x$.

Let the perturbed spatial slice be:
$ \tilde{x}_{\mathrm{in}, i} = x_{\mathrm{in}} + \varepsilon_i $
where $\varepsilon_i \sim \mathcal{N}(\mathbf{0}, \sigma^2 \mathbf{I})$, for $i \in \{1, \ldots, N\}$. 

We then construct the channel-wise perturbed input, $\tilde{x}_i$, such that the selected channel is replaced by the noisy slice, while all other channels remain unperturbed:
\[ [\tilde{x}_i]_{c} = 
\begin{cases} 
\tilde{x}_{\mathrm{in}, i} & \text{if } c = c_{\mathrm{in}} \\
[x]_c & \text{if } c \neq c_{\mathrm{in}} 
\end{cases} \]

The scalar target under this channel-specific perturbation is evaluated as:
$\mathcal{Y}_{\mathrm{target},i} = \phi\!\left([f^T(\tilde{x}_i)]_{c_{\mathrm{out}}}\right)$
and the corresponding attribution map is computed as the gradient of the target with respect to the perturbed input slice:
$G_i = \nabla_{\tilde{x}_{\mathrm{in}, i}}\,\mathcal{Y}_{\mathrm{target},i}$.

\subsection{How do structural inductive biases displace backpropagated gradients?}
\label{sec:motivation}
This section is diagnostic rather than formal. We build on established analyses of two foundational architectural paradigms in modern deep learning: Convolutional Neural Networks (CNNs)~\citep{6795724} and Transformer-based attention mechanisms~\citep{vaswani2017attention}. These two architectures exhibit fundamentally different spatial inductive biases~\citep{dosovitskiy2020image}. In both cases, we analyze how noise introduced during the forward pass propagates through the backward pass, and demonstrate that this propagation fundamentally manifests as a geometric displacement of the resulting gradients. For CNNs, we formally derive the effect via an analysis of the backward pass. For attention layers, we provide a mathematical motivation by modeling the mechanism as an interacting particle system~\citep{geshkovski2025mathematical, rigollet2025mean}, identifying architectural properties that make such geometric displacement expected. Empirical evidence is provided in Section~\ref{sec:emp_reality}. 

\subsubsection{Convolution layers}
\label{sec:noise_conv}
As detailed in Appendix~\ref{app:convolution_noise}, CNNs exhibit local geometric deformation due to the interplay of non-linearities and spatial pooling. When additive Gaussian noise $\varepsilon \sim \mathcal{N}(\bf{0}, \sigma^2 \bf{I})$ is applied to a reference input observation $x$, it first propagates linearly through the convolutional kernels $K$ during the forward pass. The resulting noise at the 
pre-activation is itself Gaussian: $\eta \sim \mathcal{N}\!\left(0,\, \tilde{\sigma}^2\right)$, where $\tilde{\sigma}^2 = \sigma^2 \|K\|_2^2$. During backpropagation, this propagated noise disrupts the backward gradient flow through two inherent phenomena:

\begin{enumerate}
    \item For a ReLU activation $y=max(0,x + \eta)$, the expected forward activation under Gaussian noise becomes $\mathbb{E}[y] = x \, \Phi\!\left(\frac{x}{\tilde{\sigma}}\right) + \tilde{\sigma} \, \phi\!\left(\frac{x}{\tilde{\sigma}}\right)$. The strictly positive term $\tilde{\sigma}\phi(x/\tilde{\sigma})$ causes dead neurons to suddenly fire. During backpropagation, these phantom activations open spurious, structurally incorrect gradient paths in the attribution map.
    \item Spatial displacement via Pooling: CNNs rely on spatial downsampling, such as Max Pooling, which route gradients strictly to the argmax coordinate within a local window. If a noise spike in a background region exceeds the true signal, the argmax operator spatially shifts. Consequently, the backpropagated gradient is physically displaced from the true object coordinate to the noise spike coordinate.
\end{enumerate}

Overall, input perturbations in CNNs inherently manifest as geometric spatial displacement in the resulting attribution maps. More technical details  are provided in Appendix~\ref{app:convolution_noise}.

\subsubsection{Attention-based layers}

\paragraph{Forward pass and attention matrix.}
Let $O_{\mathrm{in}} \in \mathbb{R}^{n \times d}$ denote 
the input token sequence to an attention layer 
and $O_{out} \in \mathbb{R}^{n \times d_v}$ its 
output. The forward pass first computes queries, keys, 
and values via learned projections 
$Q = O_{\mathrm{in}} W_Q$, $K = O_{\mathrm{in}} W_K$ and $V = O_{\mathrm{in}} W_V$. The output is then obtained via scaled dot-product attention $O_{out} = AV$, where
\begin{equation}
    A = \mathrm{softmax}\!\left(
        \frac{QK^\top}{\sqrt{d_k}}
        \right) \in \mathbb{R}^{n \times n}, 
    \label{eq:attention_forward}
\end{equation}
is the attention matrix encoding 
pairwise token similarities and  $W_Q, W_K \in \mathbb{R}^{d \times d_k}$ and $W_V \in \mathbb{R}^{d \times d_v}$ are the learned projection weights~\citep{vaswani2017attention}.

\paragraph{Gradient and dependence on attention matrix.}
As derived in Appendix~\ref{app:attention_backward}, 
the gradient with respect to 
the input sequence $O_{\mathrm{in}}$ takes the form:
\begin{equation}
    \nabla_{O_{\mathrm{in}}} \mathcal{Y}_{\mathrm{target},i} = 
A^\top \Delta_O W_V^\top + 
\frac{1}{\sqrt{D_k}} \Delta_S K W_Q^\top + 
\frac{1}{\sqrt{D_k}} \Delta_S^\top Q W_K^\top.
    \label{eq:attention_gradient}
\end{equation}

where $\Delta_O = \partial \mathcal{Y}_{\mathrm{target},i} 
/ \partial O_{out}$ is the gradient propagated 
from subsequent layers and $\Delta_S$ is the gradient 
through the softmax. The forward-pass attention 
matrix $A$  defined in Eq.~\eqref{eq:attention_forward} appears 
directly as a multiplicative factor in the 
backward pass. Therefore, the spatial structure of the 
attribution map $\nabla_{O_{\mathrm{in}}} 
\mathcal{Y}_{\mathrm{target},i}$ is directly 
anchored to the token routing induced by $A$. 
Consequently, \emph{any perturbation that alters 
$A$ directly alters the attribution map}. We show hereafter that input perturbations do precisely this.

\paragraph{Attention layers as interacting particle systems.}
To understand why input perturbations can produce changes in $A$, we adopt the interacting particle 
framework of~\citep{geshkovski2025mathematical, rigollet2025mean}, 
which models the $n$ input tokens as particles 
$\{x_i(s)\}_{i=1}^n$ evolving on the unit sphere 
$\mathbb{S}^{d-1}$ under self-attention dynamics:

\begin{equation}
\begin{cases}
\dot{x}_i(s) = \mathbf{P}_{x_i(s)}^{\perp} \left( \frac{1}{Z_{\beta,i}(s)} \sum_{j=1}^n e^{\beta \langle W_Qx_i(s), W_Kx_j(s) \rangle} W_Vx_j(s) \right), \\[6pt]
x_i(0) = x_i
\end{cases}
\label{eq:attention_ode}
\end{equation}
where $\beta > 0$ is the inverse temperature, 
$Z_{\beta,i}(s) = \sum_{k=1}^n 
e^{\beta \langle W_Q x_i(s), W_K x_k(s)\rangle}$ 
is the partition function, and 
$\mathbf{P}_{x}^{\perp} y = y - \langle x, y \rangle x$ 
is the orthogonal projection onto the tangent 
space $T_x \mathbb{S}^{d-1}$. Full derivation is given in 
Appendix~\ref{app:transformer_noise}.

In the mean-field limit $(n \rightarrow \infty)$, the empirical 
measure of the tokens $\mu_s=\frac{1}{N}\sum_{i=1}^N\delta_{x_i(s)}$ evolves according to the continuity equation 
$\partial_s \mu_s + \nabla(\mu_s \, \mathcal{X}[\mu(s)]) 
= 0$, where the vector field $\mathcal{X}[\mu(s)]$ drives the 
particles. As shown in recent literature~\citep{bruno2025a,rigollet2025mean,geshkovski2025mathematical}, this dynamics corresponds to a Wasserstein gradient flow aimed that maximizes the interaction energy $E_\beta(\mu) = \frac{1}{2\beta}\iint e^{\beta\langle x,y\rangle} 
d\mu(x)\,d\mu(y)$~\citep{geshkovski2025mathematical, rigollet2025mean} driving tokens toward semantic clusters.
The energy $E_\beta$ admits multiple local 
maxima on $\mathcal{P}(\mathbb{S}^{d-1})$, 
corresponding to qualitatively distinct 
token clustering 
configurations~\citep{rigollet2025mean, 
bruno2025a}. These multiple attractors 
are a key source of the instability we 
characterize below.

 \paragraph{Effect of input perturbation on attention routing.}
Input perturbation methods such as SmoothGrad injects additive Gaussian noise 
$\epsilon_i \sim \mathcal{N}(0, \sigma^2 I)$ 
to the input $x$. As formalized in Appendix~\ref{app:input_perturbation}, within the dynamical system framework of Eq.~\eqref{eq:attention_ode}, this is equivalent to perturbing the initial conditions: $x_i(0) = x_i + \epsilon_i$, which induces a perturbation of the initial measure $\mu_0$. Two properties of the dynamics lead to geometric displacement in the attribution maps:

First, \citep{kim2021lipschitz} demonstrated that the self-attention operator is non-Lipschitz: there is no global constant bounding the sensitivity of the output to perturbations of the input. Consequently, nearby initial conditions $x_i(0)$ and $x_i(0) + \epsilon_i$ offer no guarantee of nearby trajectories under Eq.~\eqref{eq:attention_ode}. Second, the presence of multiple local maxima in $E_\beta$ induces
metastability~\citep{rigollet2025mean, bruno2025a}, meaning that different noise realizations $\epsilon_i$ can drive the system into distinct basins of attraction, corresponding to different subsets of spatial grid coordinates that the attention clusters around.

Together, these two properties imply that the perturbed attention matrix 
$A(x + \epsilon_i)$ is not a scaled version of the clean matrix $A(x)$, 
but reflects a qualitatively different token routing configuration. 
Since the spatial structure of $\nabla_{O_{\mathrm{in}}} \mathcal{Y}_{\mathrm{target}}$ is anchored to $A$ through Eq.~\eqref{eq:attention_gradient}, this routing change propagates directly into the attribution maps. The $N$ noisy attribution maps $\{G_i\}_{i=1}^N$ are therefore not noisy estimates of a fixed attribution, but geometrically displaced versions of it. Direct empirical evidence for this claim is provided in Section~\ref{sec:emp_reality}.

\subsection{Attribution Aggregation via the Wasserstein Barycenter}
\label{sec:method}
The attribution maps $G_i$ assign an importance score 
$G_i(u) \in \mathbb{R}$ to each spatial coordinate $u \in \Omega$, where $\Omega \subset \mathbb{Z}^2$ is the spatial grid. As these scores represent output derivatives with respect to local inputs, they can be negative and possess arbitrary magnitudes, making them incompatible with OT, which operates on non-negative measures. We therefore map each $G_i$ to a discrete spatial probability distribution by taking absolute values followed by $\ell_1$ normalization:
\begin{equation}
    \mu_i(u) = \frac{|G_i(u)|}{\sum_{v \in \Omega} |G_i(v)|},
\end{equation}
where $\mu_i(u)$ represents the spatial density of feature importance at location $u$ of the input channel $c_{\mathrm{in}}$, preserving the magnitude of both positive and negative evidence, while satisfying the positivity and mass-conservation constraints of OT.
 
Our goal is now to find a single explanation that represents the geometric consensus of the displaced attribution measures $\{\mu_1, \mu_2, \dots, \mu_N\}$.
To do so, we consider $2$-Wasserstein distances $W_2(\mu, \nu)$ that account for the underlying geometry of the grid $\Omega$ by computing the minimal cost to transport mass from $\mu$ to $\nu$~\citep{alma991005863149705596}. To aggregate the $N$ samples $\mu_i$, we specifically compute the \textbf{Wasserstein Barycenter} $\mu^*$, defined as the Fréchet mean in the Wasserstein space~\citep{doi:10.1137/100805741}:
$\mu^* = \arg\min_{\mu \in \mathcal{P}(\Omega)} \sum_{i=1}^{N} W_2^2(\mu, \mu_i)$.

Since exact Wasserstein barycenter estimations are computationally prohibitive for high-dimensional grids \citep{cuturi2013sinkhorn}, we smooth the transport problem using Entropic Regularization~\citep{cuturi2013sinkhorn}:
$\mu^*_\lambda = \arg\min_{\mu\in \mathcal{P}(\Omega)} \sum_{i=1}^{N} W_{2,\lambda}^2(\mu, \mu_i)$, where $\lambda > 0$ is the regularization parameter. This formulation enables efficient computation of the barycenter via the Convolutional Sinkhorn algorithm~\citep{10.1145/2766963}. Computing $\mu^*_\lambda$ introduces a $1.96\times$ wall-time overhead compared to SmoothGrad (see Appendix~\ref{app:walltime}).

As summarized in Algorithm~\ref{alg:wasserstein_grad}, two variants based on the barycenter $\mu^*_\lambda$ are finally considered for prediction explanations on the input channel $c_{\mathrm{in}}$:
\begin{itemize}
\item
\textbf{WG\textsubscript{Bary}.}
The barycenter $\mu^*_\lambda$ can be used as the attribution map. This yields a geometrically robust explanation that captures the consensus spatial structure of the displaced measures, at the cost of losing sign information 
\item
\textbf{WG\textsubscript{Bary$\times$Grad}.}
To recover the sign of the gradient and improve the spatial concentration, $\mu^*_\lambda$ can be used as a spatial mask on the base 
gradient, so the attribution map becomes $\mu^*_\lambda \times \nabla_{x_{\mathrm{in}}} \phi\left([f^T(x)]_{c_{\mathrm{out}}}\right)$. The barycenter provides geometric localization and the base gradient provides amplitude contrast and sign.
\end{itemize}

\begin{algorithm}[t]
\caption{WassersteinGrad}
\label{alg:wasserstein_grad}
\begin{algorithmic}[1]
\REQUIRE Model $f$, input $x \in \mathbb{R}^{|\Omega|}$, number of samples $N$,
         noise variance $\sigma^2$, Sinkhorn regularization level $\lambda$, input channel of interest $c_{\mathrm{in}}$, function $\phi$ for output region/channel of interest extraction
\ENSURE  Attribution map $G_{W}$
\STATE Initialize measure list $\mathcal{M} \leftarrow [\,]$
\FOR{$i = 1$ \TO $N$}
  
    \STATE Generate noisy channel slice: $\tilde{x}_{\mathrm{in},i} \leftarrow [x]_{c_{\mathrm{in}}} + \varepsilon_i$, where $\varepsilon_i \sim \mathcal{N}(\mathbf{0}, \sigma^2 \mathbf{I})$
    \STATE Construct perturbed tensor $\tilde{x}_i$ by replacing channel $c_{\mathrm{in}}$ of $x$ with $\tilde{x}_{\mathrm{in},i}$
     \STATE Compute the perturbed prediction $f^T(\tilde{x}_i)$
    \STATE Average the prediction in a region/channel of interest $\mathcal{Y}_{\mathrm{target},i} \leftarrow \phi\left([f^T(\tilde{x}_i)]_{c_{\mathrm{out}}}\right)$ 
    \STATE Compute raw gradients $G_i\leftarrow \nabla_{\tilde{x}_{in,i}} \mathcal{Y}_{\mathrm{target},i}$
    \STATE Convert to probability measure on channel $c_{\mathrm{in}}$: $\mu_i \leftarrow \left. \dfrac{|G_i|}{\sum |G_i|}\right|_{\textrm{channel}=c_{\mathrm{in}}}$
    \STATE Append $\mu_i$ to $\mathcal{M}$
\ENDFOR
\STATE Compute $\mu^*_\lambda$, the entropic Wasserstein barycenter~\citep{10.1145/2766963} of $\mathcal{M}$ with regularization $\lambda$
\RETURN $G_W=\mu^*_\lambda$ (if WG\textsubscript{Bary}) or  $G_W=\mu^*_\lambda \times \nabla_{x_{\mathrm{in}}} \phi\left([f^T(x)]_{c_{\mathrm{out}}}\right)$ (if WG\textsubscript{Bary$\times$Grad})
\end{algorithmic}
\end{algorithm}

\section{Experiments}

\subsection{Experimental Settings}
\label{sec:setup}

\paragraph{Dataset and Model.} We evaluate on \textbf{TITAN}~\citep{meteofrance_titan_dataset}, 
a high-resolution meteorological benchmark derived from the AROME limited-area 
model at Météo-France, providing kilometre-scale ($0.025^\circ \approx 2.5$\,km) 
analyses over Western Europe. We use a subdomain over France ($512 \times 640$ 
grid points, 21 channels), with January--December 2023 as the test split. Our 
forecasting backbone is \textbf{UNetRPP}~\citep{shaker2024unetr++}, a pretrained 
hybrid convolutional--attention U-Net trained using Py4cast~\citep{meteofrance_py4cast}, 
predicting $\hat{x}_{t+1} = f(x_t)$ at 1-hour lead time with weights validated 
by meteorologists. Full details in Appendices~\ref{app:dataset}--\ref{app:model}.

\paragraph{Setup and Hyperparameters.} All methods are evaluated in a fixed-weight regime. We explain predictions of total surface precipitation (\texttt{aro\_tp\_0m}) over a Paris bounding box with respect to zonal wind at 250~hPa (\texttt{aro\_u\_250hpa}), a canonical pair reflecting the well-established jet stream influence on surface precipitation~\citep{kalnay2003atmospheric}. All stochastic methods use $N{=}20$ perturbed samples with 
$\sigma {=} 0.2 \times (\max(x_t) - \min(x_t))$, WG\textsubscript{Bary} and WG\textsubscript{Bary$\times$Grad} use Sinkhorn $\lambda{=}0.001$, selected via a faithfulness--sparsity--robustness tradeoff (Appendices~\ref{app:N_ablation}--~\ref{app:lambda_ablation}). 

\subsection{The empirical reality of Spatial Displacement}
\label{sec:emp_reality}

To visualize how attribution mass is redistributed under perturbation, we compute the 2-Wasserstein distance between the clean ($\mu_{\text{clean}}$) and perturbed ($\mu_{\text{pert}}$) gradient measures and recover the optimal transport plan $T^*$ (Appendix~\ref{app:ot_math}). By masking the diagonal of $T^*$, we isolate only the mass that physically migrated to a new spatial coordinate. As shown in Figure~\ref{fig:intro}(c), attribution mass is systematically \textit{excavated} from true feature locations (red) and \textit{deposited} into adjacent coordinates (green), forming a structured dipole pattern that is consistent with geometric displacement rather than random amplitude noise. 

Quantitatively, we measure this geometric deformation by tracking the spatial centre of mass $(c_x, c_y)$ and the peak location of the attribution maps across noise levels $\sigma \in [0.1, 1.0]$ (Appendix~\ref{app:centroid_math}). To confirm this instability is a failure mode of the attribution method rather than a degradation of the underlying forecasting model, we track this spatial displacement alongside the relative prediction error $E_{\mathrm{rel}}(\sigma) = \mathrm{RMSE}_\sigma / \mathrm{RMSE}_0$. As shown in Figure~\ref{fig:centroid_autoregressive}, the two curves dissociate sharply: below $\sigma{=}0.4$, model performance degrades by less than $1\%$, yet attribution centroids advect by an average of 10--15 km, with peak displacements exhibit even larger spatial jumps. Furthermore, this geometric displacement compounds under autoregressive rollout, reaching $\sim 115$\,km at $t{+}5$.

\subsection{Qualitative results}
\label{sec:quali_results}
\begin{figure}[h]
    \centering
    \includegraphics[width=\textwidth]{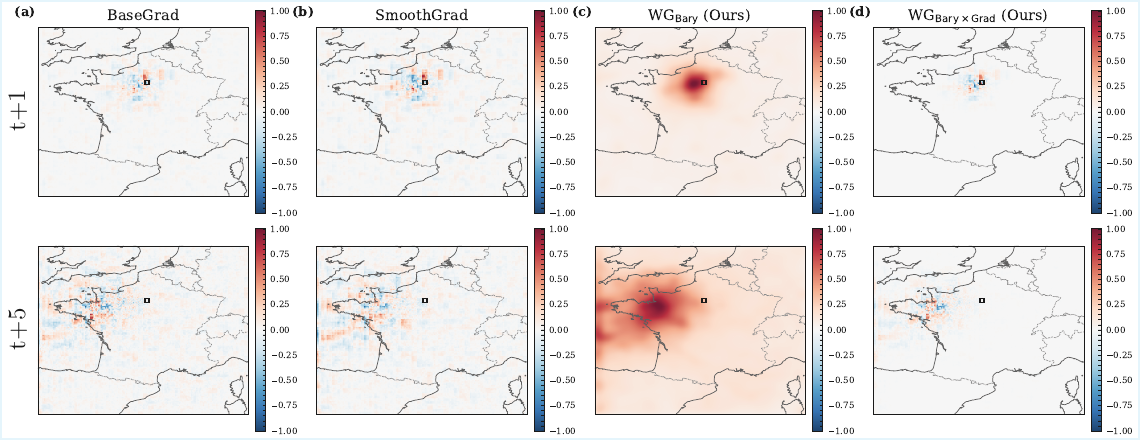} 
    \caption{Qualitative comparison of gradient attribution methods at one-step ($t+1$, top) and autoregressive ($t+5$, bottom) lead times. The attribution target is surface precipitation aggregated over the Paris region of interest (black box), with respect to input zonal wind at 250 hPa.}
    \label{fig:qualitative}
\end{figure}

\textbf{One-step forecasting.} Figure~\ref{fig:qualitative} compares attribution maps for a representative atmospheric event. \textbf{BaseGrad}~\citep{baehrens2010explain,simonyan2013deep} produces spatially scattered, high-frequency noise. \textbf{SmoothGrad}~\citep{smilkov2017smoothgrad} reduces this variance but remains spatially fragmented. In contrast, \textbf{WG\textsubscript{Bary}} yields a geographically coherent attribution localized over northern France.\textbf{WG\textsubscript{Bary$\times$Grad}} combines this geometric localization with gradient amplitude, producing a spatially concentrated map that preserves the sign of the attribution. 

\textbf{Autoregressive forecasting ($t+5$).} At longer lead times, differences between methods become more pronounced. \textbf{BaseGrad} and \textbf{SmoothGrad} remain dominated by incoherent patterns with no stable spatial organization. \textbf{WG\textsubscript{Bary}} continues to produce geographically coherent attributions, though its spatial support broadens relative to $t{+}1$ (Figure~\ref{fig:qualitative}), reflecting increased uncertainty at longer horizons. In contrast, \textbf{WG\textsubscript{Bary$\times$Grad}} yields a sparser attribution, concentrated near the Atlantic coastline, highlighting dominant contributing regions.

\subsection{Quantitative Results}
\label{sec:quant_results}
We compare our methods against four standard gradient-based baselines: \textbf{BaseGrad}~\citep{simonyan2013deep}, \textbf{SmoothGrad}~\citep{smilkov2017smoothgrad}, \textbf{VarGrad}~\citep{adebayo2018local}, and \textbf{Integrated Gradients}~\citep{sundararajan2017axiomatic}. Quantitative results, reported in Table~\ref{tab:quantitative}, are aggregated over n=100 atmospheric events from the 2023 test split characterized by high-intensity precipitation over the Paris ROI. We assess performance across both one-step (t+1) and autoregressive (t+5) lead times. Full mathematical formulations for all baseline methods and evaluation metrics (Sparsity, Faithfulness, and Robustness) are provided respectively in Appendices~\ref{app:gradient_methods} and~\ref{app:evaluation}.

\textbf{Sparsity.} WG\textsubscript{Bary} represents the geometric uncertainty of displaced attributions and is therefore less sparse than the other methods. In contrast, WG\textsubscript{Bary$\times$Grad} recovers sparsity (best at $t+5$, second-best at $t+1$) by using the barycenter as a spatial mask on the base gradient, trading geometric interpretability for sparse, sign-aware attributions.

\textbf{Faithfulness.} WG\textsubscript{Bary} achieves the highest 
faithfulness at both lead times, surpassing all baselines. 
WG\textsubscript{Bary$\times$Grad} offers competitive faithfulness with substantially higher sparsity, making it preferable when sign-aware, spatially precise attributions are required.

\textbf{Robustness.} LLE$_{\ell_2}$ measures attribution amplitude 
sensitivity, LLE$_{\cos}$ measures spatial pattern rotation under perturbation. WG\textsubscript{Bary} achieves the lowest score on \emph{both} metrics simultaneously at both lead times: a ${\sim}21{\times}$ reduction in LLE$_{\cos}$ over SmoothGrad at $t{+}1$ and degrades gracefully under autoregressive rollout, still outperforming all baselines at $t{+}5$. Notably, under autoregressive rollout, each additional forecasting step compounds geometric distortion in baseline methods, whereas WG\textsubscript{Bary}'s geometric consensus absorbs this accumulation.

\begin{table*}[t]
\caption{
    \textbf{Quantitative evaluation of attribution quality} at one-step ($t{+}1$) 
    and autoregressive ($t{+}5$) lead times. Metrics include Sparsity: Gini$\uparrow$, 
    Faithfulness: ROAD$\uparrow$, Robustness: LLE$_{\ell_2}\downarrow$ and 
    LLE$_{\cos}$$\downarrow$ ($\times 10^{3}$) on $n{=}100$ events. 
    Mean $\pm$ SEM. \textbf{Bold}: best. \underline{Underline}: second best.
}
\label{tab:quantitative}
\centering
\setlength{\tabcolsep}{3pt}
\renewcommand{\arraystretch}{0.8}
\resizebox{\textwidth}{!}{%
\begin{tabular}{l cccc cccc}
\toprule
& \multicolumn{4}{c}{\textbf{One-step} ($t+1$)} 
& \multicolumn{4}{c}{\textbf{Autoregressive} ($t+5$)} \\
\cmidrule(lr){2-5} \cmidrule(lr){6-9}
\textbf{Method} 
    & \textbf{Spar.}$\uparrow$ 
    & \textbf{Faith.}$\uparrow$ 
    & \textbf{LLE}$_{\ell_2}\downarrow$ 
    & \textbf{LLE}$_{\cos}\downarrow$
    & \textbf{Spar.}$\uparrow$ 
    & \textbf{Faith.}$\uparrow$ 
    & \textbf{LLE}$_{\ell_2}\downarrow$ 
    & \textbf{LLE}$_{\cos}\downarrow$ \\
\midrule
BaseGrad        
    & $0.547_{\pm.001}$ & $0.772_{\pm.010}$ 
    & $0.068_{\pm.001}$ & $3.5_{\pm0.07}$
    & $0.531_{\pm.002}$ & $0.714_{\pm.019}$ 
    & $0.128_{\pm.006}$ & $4.2_{\pm0.10}$ \\

IntegratedGrad        
    & $0.628_{\pm.004}$ & $\underline{0.810}_{\pm.011}$ 
    & $0.080_{\pm.001}$ & $2.9_{\pm0.10}$
    & $0.632_{\pm.004}$ & $0.739_{\pm.018}$ 
    & $0.087_{\pm.006}$ & $6.3_{\pm0.10}$ \\
SmoothGrad      
    & $0.539_{\pm.002}$ & $0.809_{\pm.012}$ 
    & $0.046_{\pm.001}$ & $\underline{2.1}_{\pm0.05}$
    & $0.526_{\pm.001}$ & $0.757_{\pm.017}$ 
    & $0.083_{\pm.004}$ & $\underline{2.6}_{\pm0.06}$ \\
VarGrad         
    & $\mathbf{0.871}_{\pm.004}$ & $0.809_{\pm.012}$ 
    & $\underline{0.024}_{\pm.001}$ & $2.3_{\pm0.06}$
    & $\underline{0.710}_{\pm.006}$ & $0.740_{\pm.013}$ 
    & $\underline{0.058}_{\pm.004}$ & $3.0_{\pm0.10}$ \\
\midrule
\textbf{WG\textsubscript{Bary}}        
    & $0.340_{\pm.002}$ & $\mathbf{0.815}_{\pm.011}$ 
    & $\mathbf{0.023}_{\pm.001}$ & $\mathbf{0.1}_{\pm0.01}$
    & $0.330_{\pm.002}$ & $\mathbf{0.762}_{\pm.016}$ 
    & $\mathbf{0.047}_{\pm.002}$ & $\mathbf{0.2}_{\pm0.01}$ \\
\textbf{WG\textsubscript{Bary$\times$Grad}}
    & $\underline{0.849}_{\pm.003}$ & $0.800_{\pm.010}$ 
    & $0.045_{\pm.001}$ & $3.6_{\pm0.08}$
    & $\mathbf{0.750}_{\pm.004}$ & $\underline{0.760}_{\pm.017}$ 
    & $0.075_{\pm.004}$ & $4.3_{\pm0.10}$ \\
\bottomrule
\end{tabular}}
\end{table*}

\section{Conclusion}

In this work, we identified a failure mode in standard gradient-smoothing XAI techniques when applied to the prediction of dynamic physical phenomena on continuous spatial domains. Through an in-depth  analysis of two common layer architectures, CNN pooling/argmax and Transformer attention as well as an empirical validation on the TITAN meteorological benchmark, we demonstrated that input perturbations do not merely induce high-frequency amplitude noise in attribution maps. Rather, they cause systematic geometric displacement of attribution mass toward physically incorrect spatial coordinates. This phenomenon is interestingly analogous to phase error in meteorological verification. We then introduced WassersteinGrad to mitigate this issue by replacing pointwise averaging in \textit{SmoothGrad} with transport-based geometric consensus via the entropic Wasserstein barycenter. Our quantitative results demonstrate that WassersteinGrad outperforms SmoothGrad and other gradient-based baselines across different metrics on meteorological data. In particular, it produces attribution maps that are geometrically coherent and physically localized.

\paragraph{Broader Impacts.} 
As deep learning increasingly replaces traditional physical solvers in high-stakes domains such as weather forecasting, the demand for reliable, physically meaningful explanations is an operational necessity. A good alignment between where a model truly attends and where its explanation reports it attends may, in practice, lead forecasters to trust neural-based predictions. We also hope this work will serve as an invitation to the community to critically reconsider the direct transfer of XAI methodology from computer vision to the physical sciences, where spatial structure and physical causality impose fundamentally different requirements on explanation quality.

\paragraph{Limitations and Future work.}

Our evaluation focused on the UNetRPP forecasting model with pre-trained parameters validated by meteorologists. Although our mechanistic analysis of backpropagated gradient displacements is applicable to any prediction model containing convolution/pooling layers or Transformer layers, its generalisation to other state-of-the-art architectures and spatial domains remains to be explored.  
A natural extension of our work concerns the noise injection strategy itself. Standard SmoothGrad and our method both perturb inputs with white Gaussian noise, which is physically uninformed. Replacing isotropic noise with physically informed perturbations could improve the physical realism of the perturbed samples. We leave this as a promising direction for future work.

\paragraph{Acknowledgements.}

This work has benefited from AI Interdisciplinary Institute ANITI. ANITI is funded by the France 2030 program under the Grant agreement n°ANR-23-IACL-0002.

\bibliographystyle{plain}
\bibliography{draft}


\appendix

\section{The TITAN Meteorological Dataset}
\label{app:dataset}

We utilize the \textbf{TITAN} dataset, a high-resolution meteorological benchmark developed by
Météo-France for deep learning-based weather forecasting~\citep{meteofrance_titan_dataset}. TITAN is derived from the operational \textbf{AROME} limited-area model, enabling
kilometre-scale forecasting over Western Europe.

\begin{figure}[t]
    \centering
    \includegraphics[width=0.82\textwidth]{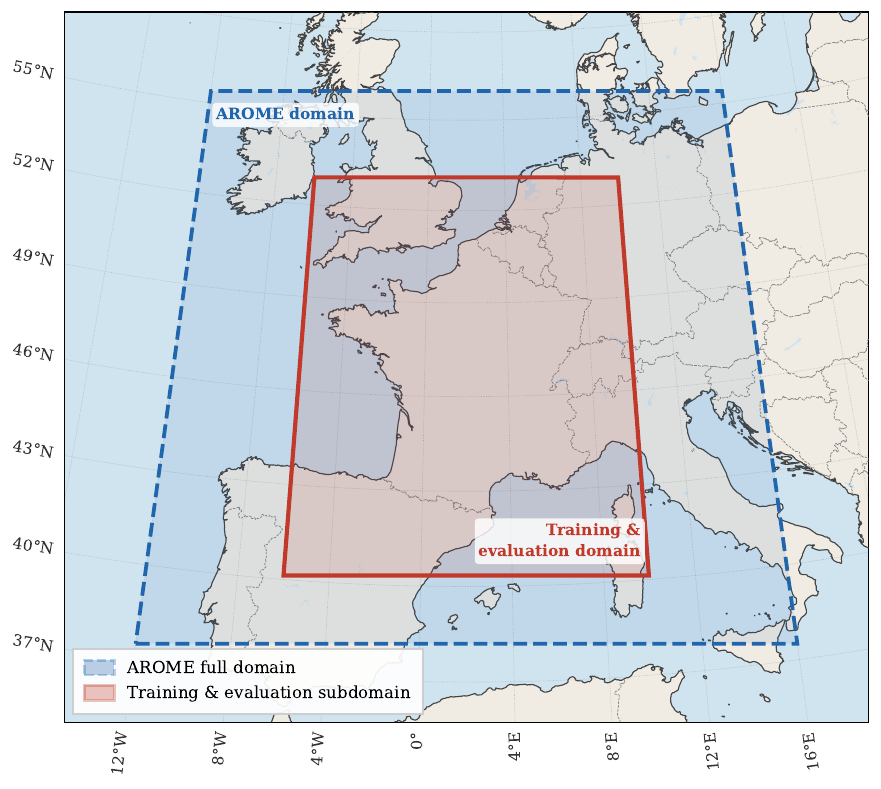}
    \caption{Geographic extent of the TITAN/AROME dataset.
             The \textcolor{blue}{blue dashed} rectangle delineates the full
             AROME operational domain; the \textcolor{red}{red solid} rectangle
             indicates the subdomain used for model training and evaluation.}
    \label{fig:titan_domain}
\end{figure}

\paragraph{Spatial Resolution.}
TITAN is projected onto a regular lat--lon grid at a native resolution of $0.025^\circ$
($\approx$\,2.5\,km).
The full AROME domain spans
$[-12^\circ,\,16^\circ] \times [37.5^\circ,\,55.4^\circ]$
($1121 \times 717$ grid points).
For all experiments we use a centred subdomain over France:
$[-6^\circ,\,9.975^\circ] \times [40.125^\circ,\,52.9^\circ]$,
corresponding to a spatial grid of $H \times W = 512 \times 640$ grid points.
Both domains are illustrated in Figure~\ref{fig:titan_domain}.

\paragraph{Temporal Resolution.}
The dataset provides analyses at a \textbf{1-hour} time step.
Our models are trained on lead times $\Delta t \in \{+1\text{h}, \ldots, +6\text{h}\}$,
with a context window of one preceding analysis step as input.
\paragraph{State Variables.}
We define the atmospheric state at timestep $t$ as a multi-channel spatial tensor
$x_t \in \mathbb{R}^{H \times W \times C}$, where $C$ is the number of
meteorological channels.
Table~\ref{tab:variables} lists all variables included in the state tensor.
\paragraph{Preprocessing and Normalization.}
Following the standard pipeline in the \texttt{py4cast} framework~\citep{meteofrance_py4cast}, all physical variables are standardized to zero mean and unit variance (Z-score normalization).

\begin{table}[t]
\caption{Meteorological variables composing the state tensor $x_t$.
         Surface diagnostics provide 5 channels; upper-air profiles at four
         pressure levels contribute 16 channels, for $C = 21$ total.}
\centering
\small
\setlength{\tabcolsep}{6pt}
\begin{tabular}{llc}
\toprule
\textbf{Variable} & \textbf{Physical Description} & \textbf{Level} \\
\midrule
\multicolumn{3}{l}{\textit{Surface and Near-Surface Variables}} \\[2pt]
\texttt{aro\_t2m} & 2-metre temperature                          & 2\,m \\
\texttt{aro\_r2}  & 2-metre relative humidity                    & 2\,m \\
\texttt{aro\_tp\_0m}  & Total accumulated precipitation              & Surface \\
\texttt{aro\_u\_10m} & 10-metre zonal wind component ($u$)          & 10\,m \\
\texttt{aro\_v\_10m} & 10-metre meridional wind component ($v$)     & 10\,m \\
\midrule
\multicolumn{3}{l}{\textit{Upper-Air Profiles (250, 500, 700, 850\,hPa)}} \\[2pt]
\texttt{aro\_t}   & Air temperature                              & $\times 4$ levels \\
\texttt{aro\_u}   & Zonal wind component ($u$)                   & $\times 4$ levels \\
\texttt{aro\_v}   & Meridional wind component ($v$)              & $\times 4$ levels \\
\texttt{aro\_z}   & Geopotential height                          & $\times 4$ levels \\
\bottomrule
\end{tabular}

\label{tab:variables}
\end{table}

\section{Model Architecture and Training Configuration}
\label{app:model}

\subsection{Architecture: UNetRPP}

We use \textbf{UNetRPP}~\citep{shaker2024unetr++}, a hybrid convolutional--attention U-Net architecture that combines a hierarchical encoder--decoder structure with multi-head self-attention at each resolution
stage. The encoder downsamples the input spatial tensor via strided
convolutions, producing a sequence of feature maps at progressively
coarser resolutions. Skip connections from each encoder stage are projected and injected into the corresponding decoder stage, following the standard U-Net architecture.
The decoder uses linear upsampling followed by a projection head.
Positional embeddings are learned via a small MLP and Instance Normalization is applied throughout.
The full architecture hyperparameters are listed in Table~\ref{tab:arch_hparams}.

\begin{table}[t]
\caption{UNetRPP architecture and forecasting hyperparameters.}

\centering
\small
\setlength{\tabcolsep}{8pt}
\begin{tabular}{lc}
\toprule
\textbf{Hyperparameter} & \textbf{Value} \\
\midrule
\multicolumn{2}{l}{\textit{Architecture}} \\[2pt]
Hidden size                          & 1024 \\
Encoder attention heads              & 16 \\
Decoder attention heads              & 4 \\
Depth per stage (encoder blocks)     & $[3,\, 3,\, 3,\, 3]$ \\
Spatial downsampling rate            & $4\times$ \\
Decoder projection size              & 64 \\
Encoder projection sizes             & $[64,\, 64,\, 64,\, 32]$ \\
Skip connections                     & \checkmark \\
Linear upsampling                    & \checkmark \\
Positional embedding                 & Learned MLP \\
Normalization                        & Instance Norm \\
Convolution operator                 & \texttt{Conv2d} \\
\midrule
\multicolumn{2}{l}{\textit{Forecasting}} \\[2pt]
Training strategy                    & Scaled auto-regressive \\
Number of auto-regressive steps      & 1 (per training iteration) \\
Loss function                        & MSE \\
\bottomrule
\end{tabular}
\label{tab:arch_hparams}
\end{table}

\subsection{Training Configuration}
\paragraph{Training strategy.}
The model was trained using the \textbf{Py4cast} framework~\citep{meteofrance_py4cast},
with the \textbf{AdamW} optimiser~\citep{loshchilov2017decoupled} ($\beta_1 = 0.9$, $\beta_2 = 0.95$), an initial learning rate of $10^{-3}$ decayed to $3 \times 10^{-7}$, and a linear warm-up of $1{,}000$ steps. Full training details including the training strategy are provided in~\citep{meteofrance_py4cast}.

\paragraph{Data Splits.}
Training split is drawn from the years 2020-2022 at 1-hour observation steps,
as summarised in Table~\ref{tab:splits}.

\begin{table}[t]
\caption{Dataset temporal splits. The validation and test sets use a 3-hour  stride between initial conditions.}
\centering
\small
\setlength{\tabcolsep}{8pt}
\begin{tabular}{lccc}
\toprule
\textbf{Split} & \textbf{Start} & \textbf{End} & \textbf{Sample step} \\
\midrule
Train      & 2020-01-01 & 2022-12-31 & 1\,h \\
Test       & 2023-01-01 & 2023-12-31 & 3\,h \\
\bottomrule
\end{tabular}
\label{tab:splits}
\end{table}
\subsection{Hyperparameter Settings}
\label{app:hyperparams}

Table~\ref{tab:hyperparams} summarises all 
hyperparameters used for attribution methods 
and evaluation metrics. All stochastic methods 
use a fixed random seed of $42$ for 
reproducibility.

\begin{table}[t]
\caption{Hyperparameter configurations. Summary of settings for the gradient-based attribution methods and parameters used for the evaluation metrics.}
\label{tab:hyperparams}
\centering
\small
\renewcommand{\arraystretch}{1.2}
\begin{tabular}{llp{5cm}}
\toprule
\textbf{Parameter} 
    & \textbf{Value} 
    & \textbf{Description} \\
\midrule
\multicolumn{3}{l}{
    \textit{
    (IntegratedGrad, SmoothGrad, VarGrad, 
    WG\textsubscript{Bary}, 
    WG\textsubscript{Bary$\times$Grad})}} \\
$N$      
    & $20$ 
    & Number of perturbed samples \\
$\sigma$ 
    & $0.2 \times (\max(x_{\mathrm{in}}) - \min(x_{\mathrm{in}}))$
    & Input perturbation noise std \\
\midrule
\multicolumn{3}{l}{
    \textit{WassersteinGrad only}} \\
$\lambda$ 
    & $0.001$ 
    & Sinkhorn entropic regularisation \\
Variant 1 
    & WG\textsubscript{Bary} 
    & Barycenter used directly as 
      attribution map \\
Variant 2 
    & WG\textsubscript{Bary$\times$Grad} 
    & Barycenter masked on base gradient 
     \\
\midrule
\multicolumn{3}{l}{
    \textit{Faithfulness (ROAD)}} \\
$p_{\max}$ 
    & $15\%$ 
    & Maximum masking percentage \\
$p$ range
    & $\{1\%, \ldots, 15\%\}$
    & Masking percentages evaluated \\
$K_{\mathrm{rand}}$ 
    & $5$ 
    & Random mask repetitions per percentage \\
Imputation
    & Linear + Gaussian noise
    & Feature removal 
      strategy~\citep{rong2022consistent} \\
$\sigma_{\mathrm{imp}}$
    & $0.1 \times (\max(x_{\mathrm{in}}) - \min(x_{\mathrm{in}}))$
    & Imputation noise std \\
\midrule
\multicolumn{3}{l}{
    \textit{Robustness ($\text{LLE}_{\ell_2}$, \text{LLE}$_{\mathrm{cosine}}$ )}} \\
$K_{\mathrm{LLE}}$ 
    & $7$ 
    & Perturbation iterations \\
$\sigma_{\mathrm{LLE}}$ 
    & $0.1 \times (\max(x_{\mathrm{in}}) - \min(x_{\mathrm{in}}))$ 
    & Perturbation noise std \\
Distance
    & $\ell_2$
    & $\| G( \mathbf{x}_{\mathrm{in}}) - G( \mathbf{x}_{\mathrm{in}} + \varepsilon) \|_2$ 
      \\
\midrule
\multicolumn{3}{l}{
    \textit{Sparsity (Gini Index)}} \\
Sorting
    & Ascending
    & Required by Gini 
      formula\\

\bottomrule
\end{tabular}
\end{table}
\section{Computational Cost and Resources}
\label{app:compute}

\subsection{Hardware and implementation}
\label{app:hardware}

All experiments were conducted on a single NVIDIA RTX A4500 Laptop GPU 
(16 GB VRAM). Implementations use PyTorch 2.5.1 with CUDA 12.1. 
Forecasting inference uses the Py4cast framework 
\citep{meteofrance_py4cast} with pretrained UNetRPP weights. Attribution 
methods, including WassersteinGrad, 
are our own implementations. For the computation of the entropic Wasserstein barycenter, we utilize the \texttt{ot.bregman.convolutional\_barycenter2d} function from the \textbf{POT} library~\citep{flamary2021pot,flamary2024pot}.
\subsection{Wall-time per sample}
\label{app:walltime}

Wall-time measurements capture the full attribution pipeline for a single sample, including gradient computation, noise sampling where applicable, and aggregation. We exclude data loading and model weight initialization from all measurements. Each method was timed over samples drawn sequentially from the 2023 test split. All methods were evaluated under identical conditions: same samples, same spatial grid resolution $512\times640$, $N=20$ perturbed samples for all stochastic methods.
Table~\ref{tab:timing} reports mean wall-time and standard deviation across 100 timed samples for each attribution method.

\begin{table}[t]
\centering
\caption{Wall-time per sample.}
\label{tab:timing}

\begin{tabular}{lccc}
\toprule
\textbf{Method} & \textbf{Mean (s)} & \textbf{Std (s)} & 
\textbf{Overhead vs.\ SmoothGrad} \\
\midrule
BaseGrad              & 0.37  & 0.03    &(no sampling) \\
SmoothGrad            & 7.42  & 0.12 & $1\times$ (reference) \\
IntegratedGrad            & 8.78  & 0.13 & $1.14\times$ \\
VarGrad               & 7.70 & 0.10 & $1.03\times$ \\
WG\textsubscript{Bary} ($\lambda = 0.01$)& 8.40 & 0.11 & $\mathbf{1.13\times}$ \\
WG\textsubscript{Bary} ($\lambda = 0.001$)& 14.60 & 0.18 & $\mathbf{1.96\times}$ \\
WG\textsubscript{Bary$\times$Grad} ($\lambda = 0.001$)   & 14.90 & 0.2 & $\mathbf{1.98\times}$ \\
\bottomrule
\end{tabular}
\end{table}

\section{Ablation Studies}
\subsection{Number of perturbed samples $N$}
\label{app:N_ablation}
We perform an ablation study on the number of perturbed inputs $N$, utilizing a fixed Sinkhorn regularization parameter of $\lambda=0.001$. Table~\ref{tab:ablation_N} reports the impact of $N$ on Sparsity, Faithfulness (ROAD), and Robustness ($\ell_2$-LLE and cosine-LLE) evaluated over $n=25$ test events. We select $N=20$ for all subsequent experiments, as it strikes a favorable balance between minimizing computational overhead and maximizing attribution stability.
\begin{table}[t]
\caption{
    \textbf{Ablation on Number of perturbed samples 
    $N$.}
    \textit{Top}: WG\textsubscript{Bary} 
    (barycenter alone).
    \textit{Bottom}: WG\textsubscript{Bary$\times$Grad} 
    (barycenter masked on base gradient).
    All metrics on $n=25$ events with highest mean precipitation in the region of interest, mean $\pm$ SEM.
    $\uparrow$ higher is better.}
\label{tab:ablation_N}
\centering
\small
\renewcommand{\arraystretch}{1.0}
\setlength{\tabcolsep}{4pt}
\begin{tabular}{lccccccc}
\toprule
$N$ 
    & \textbf{Sparsity}$\uparrow$ 
    & \textbf{Faithfulness}$\uparrow$ 
    & \textbf{LLE}$_{\ell_2}$$\downarrow$ 
    & \textbf{LLE}$_{\mathrm{cosine}}$$\downarrow$$(\times10^3)$ 
    & \textbf{Wall-time}\\
\midrule
\multicolumn{6}{l}{
    \textit{WG\textsubscript{Bary}}} \\
\midrule
$5$    
    & $0.319$\scriptsize{$\pm0.006$} 
    & $0.815$\scriptsize{$\pm0.030$}
    & $0.037$\scriptsize{$\pm0.002$}
    & $0.50$\scriptsize{$\pm0.03$} 
    & $3.70$\scriptsize{$\pm0.18$} \\

$10$   
    & $0.319$\scriptsize{$\pm0.007$} 
    & $0.825$\scriptsize{$\pm0.027$}
    & $0.032$\scriptsize{$\pm0.002$}
    & $0.41$\scriptsize{$\pm0.01$} 
    & $7.64$\scriptsize{$\pm0.20$}\\

$20$  
    & $0.320$\scriptsize{$\pm0.006$} 
    & $0.851$\scriptsize{$\pm0.023$}
    & $0.029$\scriptsize{$\pm0.003$}
    & $0.35$\scriptsize{$\pm0.03$}
    & $14.60$\scriptsize{$\pm0.20$}\\

$30$ 
    & $0.316$\scriptsize{$\pm0.006$} 
    & $0.837$\scriptsize{$\pm0.027$} 
    & $0.028$\scriptsize{$\pm0.003$} 
    & $0.34$\scriptsize{$\pm0.02$}
    & $22.51$\scriptsize{$\pm0.20$}\\
    
$40$ 
    & $0.316$\scriptsize{$\pm0.006$} 
    & $0.837$\scriptsize{$\pm0.027$} 
    & $0.027$\scriptsize{$\pm0.003$}
    &  $0.33$\scriptsize{$\pm0.02$} 
    & $32.34$\scriptsize{$\pm0.20$}\\
\midrule
\multicolumn{5}{l}{
    \textit{WG\textsubscript{Bary$\times$Grad}}} \\
\midrule
$5$    
    & $0.827$\scriptsize{$\pm0.007$} 
    & $0.840$\scriptsize{$\pm0.031$}
    & $0.042$\scriptsize{$\pm0.003$}
    & $3.64$\scriptsize{$\pm0.22$}
    & $3.96$\scriptsize{$\pm0.18$}\\

$10$   
     & $0.828$\scriptsize{$\pm0.006$} 
    & $0.821$\scriptsize{$\pm0.027$}
    & $0.042$\scriptsize{$\pm0.003$}
    & $3.63$\scriptsize{$\pm0.22$}
    & $7.70$\scriptsize{$\pm0.20$}\\

$20$  
    & $0.829$\scriptsize{$\pm0.006$} 
    & $0.824$\scriptsize{$\pm0.028$}
    & $0.042$\scriptsize{$\pm0.003$}
    & $3.63$\scriptsize{$\pm0.22$} 
    & $14.90$\scriptsize{$\pm0.20$}\\

$30$ 
    & $0.828$\scriptsize{$\pm0.006$}
    & $0.815$\scriptsize{$\pm0.029$}
    & $0.042$\scriptsize{$\pm0.003$}
    & $3.63$\scriptsize{$\pm0.20$} 
    & $22.61$\scriptsize{$\pm0.20$}\\
    
$40$ 
    & $0.828$\scriptsize{$\pm0.006$}
    & $0.825$\scriptsize{$\pm0.026$}
    & $0.042$\scriptsize{$\pm0.003$}
    & $3.63$\scriptsize{$\pm0.20$}
    & $32.46$\scriptsize{$\pm0.20$}\\
\bottomrule
\end{tabular}
\end{table}

\subsection{Sinkhorn Regularisation $\lambda$}
\label{app:lambda_ablation}

We ablate the entropic regularisation parameter 
$\lambda$ of the Sinkhorn barycenter for both 
WassersteinGrad variants. Table~\ref{tab:ablation_lambda} 
reports Sparsity, Faithfulness (ROAD), and 
Robustness ($\ell_2$-LLE and cosine-LLE) 
on $n=25$ events. We select $\lambda=0.001$ 
as the default based on the 
best faithfulness score and the sparsity-robustness tradeoff for WG\textsubscript{Bary} and WG\textsubscript{Bary$\times$Grad}, but also by the requirement that lower values of $\lambda$ provide a more accurate approximation of the true 2-Wasserstein barycenter.

\begin{table}[t]
\caption{
    \textbf{Ablation on Sinkhorn regularisation 
    $\lambda$.}
    \textit{Top}: WG\textsubscript{Bary} 
    (barycenter alone).
    \textit{Bottom}: WG\textsubscript{Bary$\times$Grad} 
    (barycenter masked on base gradient).
    All metrics on $n=25$ events with highest mean precipitation in the region of interest, mean $\pm$ SEM.
    $\uparrow$ higher is better.}
\label{tab:ablation_lambda}
\centering
\small
\renewcommand{\arraystretch}{1.0}
\setlength{\tabcolsep}{4pt}
\begin{tabular}{lcccccc}
\toprule
$\lambda$ 
    & \textbf{Sparsity}$\uparrow$ 
    & \textbf{Faithfulness}$\uparrow$ 
    & \textbf{LLE}$_{\ell_2}$$\downarrow$ 
    & \textbf{LLE}$_{\mathrm{cosine}}$$\downarrow$$(\times10^3)$\\
\midrule
\multicolumn{5}{l}{
    \textit{WG\textsubscript{Bary}}} \\
\midrule
$0.1$    
    & $0.201$\scriptsize{$\pm0.003$} 
    & $0.848$\scriptsize{$\pm0.018$}
    & $0.028$\scriptsize{$\pm0.002$}
    & $0.06$\scriptsize{$\pm0.01$} \\

$0.01$   
    & $0.287$\scriptsize{$\pm0.006$} 
    & $0.866$\scriptsize{$\pm0.024$}
    & $0.027$\scriptsize{$\pm0.002$}
    & $0.09$\scriptsize{$\pm0.01$} \\

$0.001$  
       & $0.320$\scriptsize{$\pm0.006$} 
    & $0.851$\scriptsize{$\pm0.023$}
    & $0.029$\scriptsize{$\pm0.003$}
    & $0.35$\scriptsize{$\pm0.03$} \\

$0.0005$ 
    & $0.329$\scriptsize{$\pm0.006$} 
    & $0.824$\scriptsize{$\pm0.022$}
    & $0.031$\scriptsize{$\pm0.004$}
    & $0.45$\scriptsize{$\pm0.03$} \\

\midrule
\multicolumn{5}{l}{
    \textit{WG\textsubscript{Bary$\times$Grad}}} \\
\midrule
$0.1$    
    & $0.636$\scriptsize{$\pm0.006$} 
    & $0.816$\scriptsize{$\pm0.028$}
    & $0.063$\scriptsize{$\pm0.004$}
    & $3.48$\scriptsize{$\pm0.18$} \\

$0.01$   
    & $0.767$\scriptsize{$\pm0.007$} 
    & $0.819$\scriptsize{$\pm0.032$}
    & $0.054$\scriptsize{$\pm0.003$}
    & $3.59$\scriptsize{$\pm0.19$} \\

$0.001$  
    & $0.829$\scriptsize{$\pm0.006$} 
    & $0.824$\scriptsize{$\pm0.028$}
    & $0.042$\scriptsize{$\pm0.003$}
    & $3.63$\scriptsize{$\pm0.22$} \\

$0.0005$ 
    & $0.838$\scriptsize{$\pm.006$} 
    & $0.798$\scriptsize{$\pm.035$}
    & $0.041$\scriptsize{$\pm.003$}
    & $3.72$\scriptsize{$\pm0.23$} \\
\bottomrule
\end{tabular}
\end{table}

\section{Experiment Details}
\label{app:experiment_details}

\subsection{Optimal Transport reveals spatial displacement}
\label{app:ot_math}

We model the clean and perturbed saliency maps as discrete probability 
measures over the spatial domain $\Omega \subset \mathbb{R}^2$:
\begin{equation}
    \mu_{\text{clean}}(u) = \frac{|G_{\text{clean}}(u)|}
    {\sum_{v \in \Omega} |G_{\text{clean}}(v)|}, \qquad
    \mu_{\text{pert}}(u) = \frac{|G_{\text{pert}}(u)|}
    {\sum_{v \in \Omega} |G_{\text{pert}}(v)|},
\end{equation}
where $G_{\text{clean}} = \nabla_{x_{\mathrm{in}}} \phi\left([f^T(x)]_{c_{\mathrm{out}}}\right)$ and 
$G_{\text{pert}} = \nabla_{\tilde{x}_{\mathrm{in}}} \phi\left([f^T(\tilde{x})]_{c_{\mathrm{out}}}\right)$ 
are the clean and perturbed attribution maps respectively.

The optimal transport plan $T^*$ solves:
\begin{equation}
    T^* = \arg\min_{T \in \Pi(\mu_{\text{pert}},\, \mu_{\text{clean}})} 
    \sum_{u,\, v \in \Omega} T_{uv} \cdot \|u - v\|^2,
\end{equation}
where $\Pi(\mu_{\text{pert}}, \mu_{\text{clean}})$ is the set of 
joint distributions with marginals $\mu_{\text{pert}}$ and 
$\mu_{\text{clean}}$. The marginal constraints are:
\begin{equation}
    \sum_{v \in \Omega} T_{uv} = \mu_{\text{pert}}(u), \qquad 
    \sum_{u \in \Omega} T_{uv} = \mu_{\text{clean}}(v).
\end{equation}

To isolate displaced mass, we mask the diagonal of $T^*$: entries 
$T_{uu}^*$ correspond to mass that remains at the same spatial 
coordinate $u \in \Omega$ and carry no information about 
displacement. The off-diagonal entries $T_{uv}^*$, $u \neq v$, 
encode the spatial flow of attribution mass migrating from 
coordinate $u$ to coordinate $v$ under perturbation, and are 
visualised in Figure~\ref{fig:intro}(c).

\subsection{Displacement of the center of mass}
\label{app:centroid_math}
Let $x_{\mathrm{in}}$ denote the clean input slice and let 
$G(x_{\mathrm{in}})$ denote the explanation map produced by the 
explainer. We consider additive Gaussian perturbations of the form
\begin{equation}
x_{\mathrm{in}}^{(p,r)} = x_{\mathrm{in}} + \epsilon^{(p,r)}, 
\qquad \epsilon^{(p,r)} \sim \mathcal{N}(\mathbf{0}, \sigma_p^2 \mathbf{I}),
\end{equation}
where $p$ indexes the perturbation level and $r$ the repetition. 
The noise level is defined as
\begin{equation}
\sigma_p = \alpha_p \left(\max(x_{\mathrm{in}}) - 
\min(x_{\mathrm{in}})\right),
\end{equation}
with $\alpha_p$ a scalar controlling the perturbation intensity.
For each perturbed input $x_{\mathrm{in}}^{(p,r)}$, we compute 
the corresponding explanation map:
\begin{equation}
G^{(p,r)} = \left|G\!\left(x_{\mathrm{in}}^{(p,r)}\right)\right|,
\end{equation}
and the clean explanation:
\begin{equation}
G^{(0)} = |G(x_{\mathrm{in}})|.
\end{equation}

\paragraph{Centroid of the explanation.}
The spatial centroid of the explanation map is computed using the 
centre of mass:
\begin{equation}
c_x^{(p,r)} =
\frac{\sum_{i,j} i \, G_{ij}^{(p,r)}}{\sum_{i,j} G_{ij}^{(p,r)}},
\qquad
c_y^{(p,r)} =
\frac{\sum_{i,j} j \, G_{ij}^{(p,r)}}{\sum_{i,j} G_{ij}^{(p,r)}}.
\end{equation}
Similarly, the centroid of the clean explanation is 
$(c_x^{(0)}, c_y^{(0)})$.

\paragraph{Centroid displacement.}
The displacement induced by the perturbation is measured using the 
Euclidean distance between the two centroids:
\begin{equation}
d^{(p,r)} =
\sqrt{
\left(c_x^{(p,r)} - c_x^{(0)}\right)^2 +
\left(c_y^{(p,r)} - c_y^{(0)}\right)^2
}.
\end{equation}

\paragraph{Aggregation.}
For each perturbation level $p$, we repeat the experiment $R$ times 
and compute
\begin{equation}
\bar{d}_p = \frac{1}{R} \sum_{r=1}^{R} d^{(p,r)}, \qquad
s_p = \sqrt{\frac{1}{R}\sum_{r=1}^{R} 
\left(d^{(p,r)} - \bar{d}_p\right)^2}.
\end{equation}

\paragraph{Peak displacement as complementary measure.}
We also track the spatial location of the maximum attribution, 
$(p_x, p_y) = \arg\max_{i,j} G_{ij}$, and compute its displacement 
similarly to the centroid.

\paragraph{Aggregation across events.}
For each noise level $\sigma_p$, we compute the per-event mean across 
$R$ samples, then aggregate across $n=100$ atmospheric 
events from the 2023 test split filtered by mean precipitation over 
the Paris ROI. The reported error bands are $\pm 1$ 
SEM across events.

\begin{figure}[t]
\centering
\includegraphics[width=0.9\textwidth]{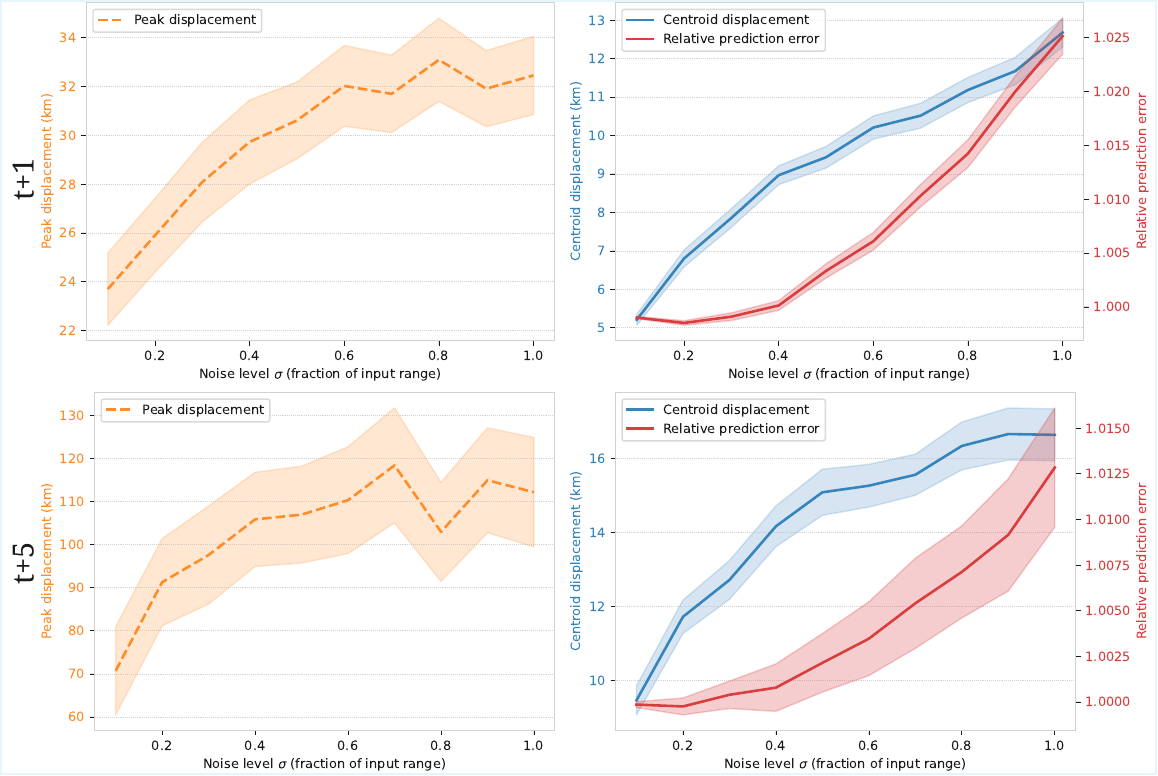}
\caption{\textbf{Centroid and peak displacement of gradient attributions 
under input perturbations, at single-step ($t+1$, top) and autoregressive 
($t+5$, bottom) lead times.} Left: peak displacement (location of 
$\arg\max|G|$). Right: centroid displacement alongside relative prediction error (secondary axis). 
Peak displacement amplifies by $3$--$4\times$ between $t+1$ and $t+5$, 
while the prediction error remains below 1.5\% in both cases. Error 
bands: $\pm 1$ SEM across $n=100$ precipitation-filtered events.}
\label{fig:centroid_autoregressive}
\end{figure}
\paragraph{Autoregressive amplification.}
Figure~\ref{fig:centroid_autoregressive} reports both centroid and peak 
displacement at single-step ($t+1$) and five-step ($t+5$) autoregressive 
horizons. The peak displacement amplifies by approximately $3$--$4\times$ 
between $t+1$ (saturating around 32\,km) and $t+5$ (reaching 115\,km), 
while the relative prediction error remains below 1.5\% at both horizons. This autoregressive amplification of geometric displacement suggests that each autoregressive step introduces 
independent geometric distortions that compound through the rollout.

\section{Gradient based methods}
\label{app:gradient_methods}
\subsection{BaseGrad}
Introduced by \citep{simonyan2013deep}, Base Gradient computes the 
partial derivatives of the output with respect to the input.
\begin{equation}
    G_{\text{Vanilla}}(x_{\mathrm{in}}) = \nabla_{x_{\mathrm{in}}} \phi\left([f^T(x)]_{c_{\mathrm{out}}}\right).
\end{equation}

\subsection{IntegratedGrad}
IntegratedGrad \citep{sundararajan2017axiomatic} addresses the local 
sensitivity of Base Gradient by integrating the gradient along a straight 
path from a baseline $x^{\mathrm{base}}$ to the input $x_{\mathrm{in}}$. In practice, the integral 
is approximated via a Riemann sum over $N$ interpolation steps $\tilde{x}_{\mathrm{in,i}} = x^{\mathrm{base}} + \frac{i}{M}(x_{\mathrm{in}} - x^{\mathrm{base}})$:
\begin{equation}
    G_{\text{IG}}(x_{\mathrm{in}}) = \frac{x_{\mathrm{in}} - x^{\mathrm{base}}}{N} 
    \sum_{I=1}^{N} \nabla_{\tilde{x}_{\mathrm{in,i}}}\phi\left([f^T(\tilde{x}_{i})]_{c_{\mathrm{out}}}\right).
\end{equation}
We define $x^{\mathrm{base}}$ as the climatological mean of the training set, representing a neutral atmospheric state. Since our inputs are Z-standardized, this baseline corresponds to a neutral atmospheric state where $x^{\mathrm{base}} = 0$.
\subsection{SmoothGrad}
To mitigate local gradient shattering, SmoothGrad 
\citep{smilkov2017smoothgrad} computes the expected gradient over a 
local Gaussian neighbourhood, approximated via Monte Carlo sampling 
over $N$ perturbed inputs $\tilde{x}_{\mathrm{in},i} = x_{\mathrm{in}} + \varepsilon_i$, where 
$\varepsilon_i \sim \mathcal{N}(\mathbf{0}, \sigma^2 \mathbf{I})$:
\begin{equation}
    G_{\text{SG}}(x_{\mathrm{in}}) = \frac{1}{N} \sum_{i=1}^N 
    \nabla_{\tilde{x}_{\mathrm{in,i}}}\, \phi([f^T(\tilde{x}_{i})]_{c_{\mathrm{out}}}).
\end{equation}

\subsection{VarGrad}
VarGrad \citep{adebayo2018local} replaces the mean of SmoothGrad 
with the variance of the perturbed gradients, highlighting regions 
where the gradient is highly sensitive to local noise:
\begin{equation}
    G_{\text{Var}}(x_{\mathrm{in}}) = \frac{1}{N} \sum_{i=1}^N 
    \left(\nabla_{\tilde{x}_{\mathrm{in,i}}}\, \phi([f^T(\tilde{x}_{i})]_{c_{\mathrm{out}}}) - 
    G_{\text{SG}}(x_{\mathrm{in}})\right)^2.
\end{equation}

\section{Evaluation Metrics}
\label{app:evaluation}

To evaluate the quality of our attribution maps within the continuous 
spatial domain of weather forecasting, we adopt a suite of 
quantitative metrics inspired by recent XAI 
benchmarks~\citep{bommer2024finding}. We evaluate explanations across 
three fundamental axes: Faithfulness, Robustness, and Complexity.

\paragraph{Faithfulness (ROAD-MSE $\uparrow$).}
Features marked as highly salient must strictly influence the 
model's prediction. We evaluate this using the RemOve And Debias 
(ROAD) framework~\citep{rong2022consistent}. Because weather 
forecasting is a regression task, raw RMSE degradations 
vary by orders of magnitude across events , making cross-event aggregation of continuous scores unstable. We therefore adapt ROAD to this setting via 
a binary comparison restricted to the region of interest 
$\mathcal{B}$. For each masking percentage $p \in \{1\%, 2\%, 
\ldots, p_{\text{max}}\}$, we mask the top-$p\%$ salient pixels 
using linear imputation~\citep{rong2022consistent} and compare the 
resulting prediction degradation against the mean of $K$ random 
masks of identical size:
\begin{equation}
    \mathrm{binary}(p) = 
    \mathbf{1}\!\left[
        \left|\phi\!\left([f^T(x)]_{c_{\mathrm{out}}}\right) - 
        \phi\!\left([f^T(x^{\mathrm{sal},p})]_{c_{\mathrm{out}}}\right)\right|
        > \frac{1}{K}\sum_{k=1}^K
        \left|\phi\!\left([f^T(x)]_{c_{\mathrm{out}}}\right) - 
        \phi\!\left([f^T(x^{\mathrm{rand},p,k})]_{c_{\mathrm{out}}}\right)\right|
    \right].
    \label{eq:binary_road}
\end{equation}
where $x^{\mathrm{sal},p}$ denotes the input with top-$p\%$ salient 
pixels masked, $x^{\mathrm{rand},p,k}$ denotes the $k$-th random 
mask of the same size, and $\phi$ is the ROI aggregation operator 
defined in Section~\ref{sec:pipeline}. The faithfulness score is 
the AUC of the binary curve:
\begin{equation}
    q_{\mathrm{ROAD}} = 
    \frac{\int_1^{p_{\text{max}}} \mathrm{binary}(p)\,\mathrm{d}p}
    {p_{\text{max}} - p_{\text{min}}}.
    \label{eq:road_auc}
\end{equation}

\paragraph{Robustness ($\text{LLE}_{\ell_2}$ $\downarrow$).}
A reliable explanation must remain stable under noise. We quantify 
this using the Local Lipschitz Estimate 
(LLE)~\citep{alvarez2018robustness}, which measures the worst-case 
instability of the attribution map $G$ under a bounded input 
perturbation:
\begin{equation}
    q_{\text{LLE}_{\ell_2}} = \max_{\|\varepsilon\|_2 \leq \delta} 
    \frac{\| G(x_{\mathrm{in}}) - 
    G(x_{\mathrm{in}} + \varepsilon) \|_2}{\|\varepsilon\|_2},
\end{equation}

A lower score indicates greater robustness to local perturbations.

\paragraph{Robustness ($\text{LLE}_{\mathrm{cosine}}$ $\downarrow$).}
To measure \emph{structural} rather than amplitude change, we 
normalize the attribution maps before comparison:
\begin{equation}
    q_{\text{LLE}_{\mathrm{cosine}}} = 
    \max_{\|\varepsilon\|_2 \leq \delta} 
    \frac{\| \hat{G}(x_{\mathrm{in}}) - 
    \hat{G}(x_{\mathrm{in}} + \varepsilon) \|_2}
    {\|\varepsilon\|_2},
\end{equation}
where $\hat{G} = G / \|G\|_2$ denotes the unit-normalized 
attribution map. This metric penalises methods whose attribution structure changes under perturbation.

\paragraph{Complexity (Gini Index $\uparrow$).}
A highly distributed explanation is often uninterpretable for domain 
experts. We quantify spatial concentration using the Gini 
Index~\citep{chalasani2020concise}:
\begin{equation}
    q_{\text{SPA}} = \frac{\sum_{i=1}^d (2i - d - 1)\, 
    |G(x_{\mathrm{in}})_{(i)}|}{d \sum_{i=1}^d 
    |G(x_{\mathrm{in}})_{(i)}|},
\end{equation}
where $d = H \times W$ is the total number of spatial grid points 
and $|G(x_{\mathrm{in}})_{(i)}|$ denotes the $i$-th smallest absolute attribution 
value. A higher Gini coefficient indicates a sparser, more localised 
explanation.

\section{Mathematical Preliminaries}
\subsection{Optimal Transport and Wasserstein Barycenters}
\label{app:optimal_transport}

\subsubsection{The 2-Wasserstein Distance}

Let $\mu, \nu \in \mathcal{P}(\Omega)$ be two probability measures 
defined on the spatial grid $\Omega$. The $2$-Wasserstein distance 
is defined via the Kantorovich formulation:
\begin{equation}
    W_2^2(\mu, \nu) = \min_{\pi \in \Pi(\mu, \nu)} 
    \int_{\Omega \times \Omega} \|u - v\|_2^2 \, d\pi(u, v),
\end{equation}
where $\Pi(\mu, \nu)$ denotes the set of all joint probability 
couplings with marginals $\mu$ and $\nu$.

\subsubsection{Discrete 2-Wasserstein Distance}

In our setting, spatial explanations are defined on a discrete grid 
of $n$ locations. We represent them as normalized non-negative 
vectors $\mu, \nu \in \Delta^n$, where
\begin{equation}
    \Delta^n = \left\{ p \in \mathbb{R}_+^n \;\middle|\; 
    \sum_{i=1}^n p_i = 1 \right\}.
\end{equation}
Let $M \in \mathbb{R}^{n \times n}$ be the cost matrix, where 
$M_{ij} = \|u_i - u_j\|_2^2$ encodes the squared Euclidean 
distance between spatial coordinates $u_i, u_j \in \Omega$. The 
discrete $2$-Wasserstein distance is:
\begin{equation}
    W_2^2(\mu, \nu) = \min_{\pi \in \mathbb{R}_+^{n \times n}} 
    \langle M, \pi \rangle \quad \text{s.t.} \quad 
    \pi \mathbf{1} = \mu, \quad \pi^\top \mathbf{1} = \nu,
    \label{eq:w2_discrete}
\end{equation}
where $\pi$ is a transport plan and 
$\langle M, \pi \rangle = \sum_{i,j} M_{ij}\pi_{ij}$. This linear 
program scales as $\mathcal{O}(n^2)$ variables and is prohibitive 
for high-resolution grids.

\subsubsection{Entropic Regularization and Sinkhorn Iterations}

To obtain a tractable formulation, we use entropic 
regularization~\citep{cuturi2013sinkhorn}:
\begin{equation}
    W_{2,\lambda}^2(\mu,\nu) = \min_{\pi} \langle M, \pi \rangle 
    + \lambda \sum_{i,j} \pi_{ij}(\log \pi_{ij} - 1),
    \label{eq:entropic_ot}
\end{equation}
subject to the same marginal constraints. The entropy term makes 
the objective strictly convex and yields the scaling form
\begin{equation}
    \pi^\star = \mathrm{diag}(a)\, \mathbf{K}_\lambda\, 
    \mathrm{diag}(b), \qquad 
    \mathbf{K}_\lambda = \exp(-M/\lambda),
\end{equation}
where the scaling vectors $a$ and $b$ are computed via alternating 
matrix-vector normalizations (Sinkhorn iterations), which are highly 
parallelisable on GPUs~\citep{cuturi2013sinkhorn}.

\subsection{Mean-Field Limits and Gradient Flows}

\subsubsection{The Mean-Field Regime}

The \textit{mean-field regime} approximates the behaviour of a 
large number of interacting particles ($n \to \infty$) by tracking 
their statistical distribution rather than individual trajectories.

Consider a discrete system of $n$ particles (tokens) 
$\{x_i(s)\}_{i=1}^n$ evolving according to a coupled ODE:
\begin{equation}
    \dot{x}_i(s) = \frac{1}{n} \sum_{j=1}^n 
    \mathcal{K}(x_i(s), x_j(s)),
\end{equation}
where $\mathcal{K}: \mathbb{R}^D \times \mathbb{R}^D \to 
\mathbb{R}^D$ is an interaction kernel (\textit{e.g.}\ the 
attention mechanism) and $D$ is the token embedding dimension. 
We introduce the \textbf{empirical measure} $\mu_n(s)$, a discrete 
probability measure describing the state of the system at time $s$:
\begin{equation}
    \mu_n(s) = \frac{1}{n} \sum_{i=1}^n \delta_{x_i(s)},
\end{equation}
where $\delta_x$ denotes the Dirac mass at position $x$.

\paragraph{The Mean-Field Limit ($n \to \infty$).}
As $n \to \infty$, the empirical measure $\mu_n(s)$ converges 
weakly to a deterministic continuous probability measure $\mu_s$. 
The evolution of this limiting measure is governed by the 
\textbf{continuity equation}:
\begin{equation}
    \partial_s \mu_s + \nabla \cdot 
    (\mu_s\, v_s[\mu_s]) = 0,
\end{equation}
where the velocity field $v_s[\mu_s]$ at a point $x$ depends on 
the entire distribution $\mu_s$ via convolution with the kernel:
\begin{equation}
    v_s[\mu_s](x) = \int \mathcal{K}(x, y) \, d\mu_s(y).
\end{equation}
This equation describes the transport of mass: the local change in 
density $\partial_s \mu_s$ is fully accounted for by the divergence 
of the flux $\mu_s v_s$.

\subsubsection{Wasserstein Gradient Flows}

A \textit{gradient flow} describes the evolution of a system 
following the path of steepest descent (or ascent) to minimise 
(or maximise) an energy functional. In the space of probability 
measures, the geometry is defined by the \textbf{Wasserstein 
metric}.

\paragraph{The Gradient Flow Structure.}
We consider an energy functional 
$\mathcal{E}: \mathcal{P}(\mathbb{R}^D) \to \mathbb{R}$. In the 
context of attention, this is the \textbf{interaction energy}:
\begin{equation}
    \mathcal{E}_{\beta}(\mu) = \frac{1}{2\beta} 
    \iint e^{\beta \langle x, y \rangle} \, d\mu(x) \, d\mu(y).
\end{equation}
A curve of measures $(\mu_s)_{s \ge 0}$ is a \textbf{Wasserstein 
gradient flow} for $\mathcal{E}$ if it satisfies the continuity 
equation with velocity field equal to the negative Wasserstein 
gradient of the energy:
\begin{equation}
    v_s = -\nabla_{W_2} \mathcal{E}(\mu_s) = 
    -\nabla \left( \frac{\delta \mathcal{E}}{\delta \mu} \right),
\end{equation}
where $\frac{\delta \mathcal{E}}{\delta \mu}$ denotes the first 
variation of the energy.

\section{Noise Effect in Convolutional Architectures}
\label{app:convolution_noise}

In Section~\ref{sec:noise_conv}, we established that input 
perturbations in CNNs manifest as spatial geometric displacement 
in the resulting attribution maps. In this section, we formally 
derive this phenomenon by tracing the propagation of additive 
Gaussian noise through the forward pass (convolutions, 
non-linearities, and pooling) and analysing its impact on the 
backward pass.

\subsection{Gradient Path}

We analyse the backward pass using the path formulation of the 
gradient, following~\citep{hill2025smoothed,balduzzi2017shattered}. 
Applying the multivariable chain rule along the computational graph, 
the partial derivative of the output 
$\mathcal{Y}_{\mathrm{target}} = f^T(x)$ with respect to the 
$i$-th input feature $x_i$ is the sum of derivatives along all 
paths connecting them:
\begin{equation}
    \left[ \nabla f^T(x) \right]_i 
    = \frac{\partial f^T(x)}{\partial x_i}
    = \sum_{\rho \in P_{i \to c}} \prod_{n \in \rho} 
    \left[ J_{f^n}(\alpha^n) \right]_\rho,
\end{equation}
where $\alpha^n$ denotes the input activations to function $f^n$.

\paragraph{Factorization of Linear and Nonlinear Contributions.}
Following~\citep{hill2025smoothed}, we factorize the terms in 
this path product into two categories:
\begin{enumerate}
    \item \textbf{Linear or affine functions} (\textit{e.g.}\ dense 
    and convolutional layers): their Jacobian entries are constant 
    weights,
    \[
        \left[ J_{f^n} \right]_\rho = W_\rho^n.
    \]
    \item \textbf{Nonlinear functions} (\textit{e.g.}\ ReLU, max 
    pooling): their Jacobian entries depend on the input activation 
    $\alpha^n$,
    \[
        \left[ J_{f^n}(\alpha^n) \right]_\rho = 
        g_\rho^n(\alpha^n).
    \]
\end{enumerate}
Separating the constant linear weights from the input-dependent 
nonlinear derivatives yields:
\begin{equation}
    \frac{\partial f^T(x)}{\partial x_i}
    = \sum_{\rho \in P_{i \to c}}
    \left( \prod_{u \in \rho} W_\rho^u \right)
    \left( \prod_{v \in \rho} g_\rho^v(\alpha^v) \right).
\end{equation}
We collapse these products for notational clarity:
\begin{align*}
    W_\rho &= \prod_{u \in \rho} W_\rho^u 
    \quad \text{(path-weight)}, \\
    \Gamma_\rho(x) &= \prod_{v \in \rho} g_\rho^v(\alpha^v)
    \quad \text{(path-indicator function)},
\end{align*}
giving:
\begin{equation}
    \frac{\partial f^T(x)}{\partial x_i}
    = \sum_{\rho \in P_{i \to c}} W_\rho\, \Gamma_\rho(x).
    \label{eq:path_gradient}
\end{equation}
Previous works view the fragility of $\Gamma_\rho(x)$ under input 
perturbations through the lens of the Shattered Gradient 
Problem~\citep{hill2025smoothed,balduzzi2017shattered}.

\subsection{Linear Propagation in Convolutions}

Consider a standard 2D convolutional layer $l$. Let 
$o^l \in \mathbb{R}^{H \times W}$ denote the input feature map 
and $\mathbf{K}^l \in \mathbb{R}^{(2V+1) \times (2U+1)}$ denote 
the convolutional kernel. The clean pre-activation $z^{l+1}_{h,w}$ 
at spatial location $(h, w)$ is:
\begin{equation}
    z^{l+1}_{h,w} = \sum_{m=-V}^{V} \sum_{n=-U}^{U} 
    \mathbf{K}^{l}_{m,n}\, o^{l}_{h+m,\, w+n}.
\end{equation}
We assume the input to layer $l$ is corrupted by i.i.d.\ additive 
Gaussian noise:
\begin{equation}
    \tilde{o}^l_{h,w} = o^l_{h,w} + \varepsilon^l_{h,w}, 
    \qquad \varepsilon^l_{h,w} \sim \mathcal{N}(0,\sigma^2).
\end{equation}
Substituting into the convolution yields:
\begin{equation}
    \tilde{z}^{l+1}_{h,w} = 
    \underbrace{\sum_{m,n} \mathbf{K}^l_{m,n}\, 
    o^l_{h+m,\,w+n}}_{\text{clean signal } z^{l+1}_{h,w}} 
    + \underbrace{\sum_{m,n} \mathbf{K}^l_{m,n}\, 
    \varepsilon^l_{h+m,\,w+n}}_{\text{noise term } 
    \eta^{l+1}_{h,w}}.
\end{equation}
Because the noise variables are independent Gaussians and the 
convolution is linear, the noise term $\eta^{l+1}_{h,w}$ remains 
Gaussian with variance scaled by the squared $\ell_2$ norm of the 
kernel:
\begin{equation}
    \eta^{l+1}_{h,w} \sim \mathcal{N}\!\left(0,\, \sigma^2 
    \|\mathbf{K}^l\|_2^2\right), \qquad 
    \|\mathbf{K}^l\|_2^2 = \sum_{m=-V}^{V}\sum_{n=-U}^{U} 
    \left(\mathbf{K}^l_{m,n}\right)^2.
\end{equation}
Thus $\tilde{z}^{l+1}_{h,w} = z^{l+1}_{h,w} + \eta^{l+1}_{h,w}$. 
The noise propagates linearly until it interacts with the 
non-linear activation $\varphi$.

\subsection{Impact on Smooth Activations (General Case)}

For smooth, twice-differentiable activation functions 
(\textit{e.g.}\ GELU, SiLU), we expand via a second-order Taylor 
series around the clean pre-activation $z^{l+1}_{h,w}$:
\begin{align}
    \tilde{o}^{l+1}_{h,w} 
    &= \varphi\!\left(z^{l+1}_{h,w} + \eta^{l+1}_{h,w}\right) 
    \nonumber \\
    &= \varphi(z^{l+1}_{h,w}) 
    + \varphi'(z^{l+1}_{h,w})\,\eta^{l+1}_{h,w} 
    + \tfrac{1}{2}\varphi''(z^{l+1}_{h,w})
    \left(\eta^{l+1}_{h,w}\right)^2 
    + \mathcal{O}\!\left((\eta^{l+1}_{h,w})^3\right).
\end{align}
Taking the expectation and using $\mathbb{E}[\eta]=0$ and 
$\mathbb{E}[\eta^2] = \sigma^2\|\mathbf{K}^l\|_2^2$:
\begin{equation}
    \mathbb{E}\big[\tilde{o}^{l+1}_{h,w}\big] = 
    \varphi(z^{l+1}_{h,w}) + \frac{1}{2}\,
    \varphi''(z^{l+1}_{h,w})\,\sigma^2 
    \|\mathbf{K}^l\|_2^2 + \mathcal{O}(\sigma^3).
\end{equation}
The expected activation is shifted from the clean activation 
proportionally to the propagated noise variance and the local 
curvature $\varphi''$.

\subsection{Analysis for Piecewise Linear Activations (ReLU)}

For piecewise linear functions such as ReLU, the second derivative is zero almost everywhere, rendering the Taylor expansion uninformative. Let $z$ denote the clean pre-activation and $\eta \sim \mathcal{N}(0, \tilde{\sigma}^2)$ the propagated noise.

We compute the expected value:
\[
\mathbb{E}[y]
=
\mathbb{E}[ReLu(z+\eta)]
=
\mathbb{E}[\max(0,z+\eta)].
\]

Using the Gaussian density
\[
\phi_{\tilde{\sigma}}(\varepsilon)
=
\frac{1}{\sqrt{2\pi}\tilde{\sigma}}
\exp\!\left(-\frac{\varepsilon^2}{2\tilde{\sigma}^2}\right),
\]
we integrate:

\begin{align*}
   \mathbb{E}[y] &=
\int_{-\infty}^{\infty}
\max(0,z+\eta)
\, \phi_{\tilde{\sigma}}(\eta)
\, d\eta \\
&= \int_{-z}^{\infty} (z+\eta) \, \phi_{\tilde{\sigma}}(\eta) \, d\eta \\
&=  z \, \Phi\!\left(\frac{z}{\tilde{\sigma}}\right) + \tilde{\sigma} \, \phi\!\left(\frac{z}{\tilde{\sigma}}\right) 
\end{align*}

where

\begin{align*}
    \Phi(u)
&=
\int_{-\infty}^{u}
\frac{1}{\sqrt{2\pi}}
e^{-t^2/2}
\, dt
\quad\text{(Gaussian CDF)} \\
\phi(u)
&=
\frac{1}{\sqrt{2\pi}}
e^{-u^2/2}
\quad\text{(Gaussian PDF)}
\end{align*}

\textbf{Creation of Phantom Gradients:} Even when the clean pre-activation is negative ($z < 0$) and the neuron should be strictly inactive, the term $\tilde{\sigma} \phi\!\left(z/\tilde{\sigma}\right)$ remains strictly positive. Consequently, the injected noise creates \textit{phantom activations}: neurons that would be strictly inactive in the clean pass now exhibit a positive expected output. During the backward pass of attribution generation, these phantom activations prematurely unblock gradient flow through the ReLU gates, opening spurious attribution paths $\Gamma_\rho(x)$ in~\eqref{eq:path_gradient} that highlight regions entirely irrelevant to the original feature geometry~\citep{hill2025smoothed}.

\section{Backpropagation and Noise Effect in Attention Layers}
\label{app:attention_backward}

\paragraph{Setup in the XAI context.}
In our attribution framework, the scalar target 
$\mathcal{Y}_{\mathrm{target}}$ is the regional forecast defined 
in Section~\ref{sec:pipeline}. The attribution map 
$\nabla_{x_{\mathrm{in}}}\mathcal{Y}_{\mathrm{target}}$ measures 
the sensitivity of this scalar forecast to the input channel 
$x_{\mathrm{in}}$. Within a Transformer attention layer, the input 
token sequence is $O_{\mathrm{in}} \in \mathbb{R}^{n \times D}$ 
(the feature map at that stage, with $n$ spatial positions 
and embedding dimension $D$), and 
$O_{\mathrm{out}} \in \mathbb{R}^{n \times D_v}$ is its output. 
We derive 
$\nabla_{O_{\mathrm{in}}} \mathcal{Y}_{\mathrm{target}}$ by 
backpropagating through the attention mechanism.

\paragraph{Forward pass.}
Queries, keys, and values are computed via learned projections 
$W_Q, W_K \in \mathbb{R}^{D \times D_k}$, 
$W_V \in \mathbb{R}^{D \times D_v}$:

\begin{equation}
    Q = O_{\mathrm{in}} W_Q, \quad 
    K = O_{\mathrm{in}} W_K, \quad 
    V = O_{\mathrm{in}} W_V.
    \label{eq:qkv}
\end{equation}
Scaled dot-product attention yields:
\begin{equation}
    S = \frac{QK^\top}{\sqrt{D_k}}, 
    \qquad
    A = \mathrm{softmax}(S) \in \mathbb{R}^{n \times n},
    \qquad
    O_{\mathrm{out}} = AV,
    \label{eq:attention_fwd}
\end{equation}
where softmax is applied row-wise: 
$A_{ij} = \exp(S_{ij}) / \sum_{j'}\exp(S_{ij'})$.

\paragraph{Gradients with respect to projections.}
Let $\Delta_O = \partial\mathcal{Y}_{\mathrm{target}} / 
\partial O_{\mathrm{out}} \in \mathbb{R}^{n \times D_v}$ denote 
the upstream gradient propagated from subsequent layers. By the 
chain rule:
\begin{align}
    \frac{\partial \mathcal{Y}_{\mathrm{target}}}{\partial V} 
    &= A^\top \Delta_O, 
    \label{eq:grad_V} \\[4pt]
    \Delta_A = \frac{\partial \mathcal{Y}_{\mathrm{target}}}
    {\partial A} &= \Delta_O V^\top.
    \label{eq:grad_A}
\end{align}
Defining $\Delta_S = \partial\mathcal{Y}_{\mathrm{target}} 
/ \partial S$:
\begin{align}
    \frac{\partial \mathcal{Y}_{\mathrm{target}}}{\partial Q} 
    &= \frac{1}{\sqrt{D_k}} \Delta_S K, 
    \label{eq:grad_Q} \\[4pt]
    \frac{\partial \mathcal{Y}_{\mathrm{target}}}{\partial K} 
    &= \frac{1}{\sqrt{D_k}} \Delta_S^\top Q.
    \label{eq:grad_K}
\end{align}

\paragraph{Gradient through the softmax.}
The row-wise softmax Jacobian gives, for row $i$:
\begin{equation}
    (\Delta_S)_{ij} = A_{ij}\!\left(
        (\Delta_A)_{ij} - \sum_k A_{ik}(\Delta_A)_{ik}
    \right).
    \label{eq:grad_softmax_row}
\end{equation}
In matrix form via the Hadamard product:
\begin{equation}
    \Delta_S = A \odot \!\left(
        \Delta_A - (\Delta_A \odot A)
        \mathbf{1}\mathbf{1}^\top
    \right),
    \label{eq:grad_softmax}
\end{equation}
where $\mathbf{1} \in \mathbb{R}^n$ is a vector of ones, making 
$(\Delta_A \odot A)\mathbf{1}\mathbf{1}^\top$ a matrix whose 
columns equal the row-sums of $\Delta_A \odot A$.

\paragraph{Total gradient with respect to the input.}
Applying the chain rule backward through 
Eqs.~\eqref{eq:qkv}--\eqref{eq:grad_K}:
\begin{equation}
\label{eq:total_attention_grad}
\boxed{
\nabla_{O_{\mathrm{in}}} \mathcal{Y}_{\mathrm{target}} = 
A^\top \Delta_O W_V^\top + 
\frac{1}{\sqrt{D_k}} \Delta_S K W_Q^\top + 
\frac{1}{\sqrt{D_k}} \Delta_S^\top Q W_K^\top.
}
\end{equation}
\section{The Mean-Field Dynamics of Transformers}
\label{app:transformer_noise}

The Transformer architecture has become the dominant framework in 
modern deep learning. While many variants exist to stabilize the 
architecture, the core mathematical interest lies in understanding 
how the attention mechanism itself regulates token 
dynamics~\citep{rigollet2025mean}.

\subsection{Discrete to Continuous Time}
 We denote autoregressive forecasting steps as $t$, and the continuous-time depth within a single model pass as $s$.
 
Transformers process data through a sequence of discrete layers 
$k = 1, \dots, L$. This recursive structure naturally suggests a 
discrete-time dynamical system. As the depth increases, we view 
the layer index as a continuous variable $s$, transitioning from 
discrete updates to a continuous-time nonlinear flow between 
tokens~\citep{rigollet2025mean,geshkovski2025mathematical}:
\begin{equation}
    \dot{x}(s) = F_s(x(s)).
\end{equation}
The attention mechanism defines a \textbf{nonlocal velocity field}, 
coupling each token to all others through a kernel that depends on 
their pairwise similarities.

\subsection{The Dynamics of Self-Attention}

We model the $n$ tokens as particles $x_i(s)$ evolving on the unit 
sphere $\mathbb{S}^{D-1}$.

\subsubsection{Standard Self-Attention (SA)}

The dynamics of the standard self-attention mechanism, 
incorporating the geometric constraint of layer normalization 
(keeping particles on the sphere), are given 
by~\citep{rigollet2025mean,geshkovski2025mathematical}:
\begin{equation}
    \dot{x}_i(s) = \mathbf{P}_{x_i(s)}^{\perp} \!\left( 
    \frac{1}{Z_{\beta,i}(s)} \sum_{j=1}^n 
    e^{\beta \langle W_Q x_i(s),\, W_K x_j(s) \rangle} 
    W_V x_j(s) \right),
\end{equation}
where:
\begin{itemize}
    \item $\beta > 0$ is the inverse temperature parameter,
    \item $Z_{\beta,i}(s) = \sum_{\ell=1}^n 
    e^{\beta \langle x_i(s), x_\ell(s) \rangle}$ is the partition 
    function,
    \item $\mathbf{P}_{x}^{\perp} y = y - \langle x, y \rangle x$ 
    is the orthogonal projection onto the tangent space 
    $T_x \mathbb{S}^{D-1}$.
\end{itemize}
For simplicity we take $W_V = W_Q = W_K = \mathrm{Id}$. This 
equation describes $n$ particles interacting through the kernel 
$\mathcal{K}(x, y) = e^{\beta \langle x, y \rangle}$.

\subsubsection{Unnormalized Self-Attention (USA)}

To simplify analysis, we consider a variant that omits the 
normalization denominator $Z$. The unnormalized self-attention 
dynamics are:
\begin{equation}
\begin{cases}
    \dot{x}_i(s) = \mathbf{P}_{x_i(s)}^{\perp} \!\left( 
    \dfrac{1}{n} \sum_{j=1}^n 
    e^{\beta \langle x_i(s), x_j(s) \rangle} x_j(s) 
    \right), \\[8pt]
    x_i(0) = x_i.
\end{cases}
\end{equation}
In practice, the behaviour of USA closely mirrors that of SA, 
making it a valuable proxy for theoretical 
study~\citep{rigollet2025mean,geshkovski2025mathematical}.

\subsection{Gradient Flows and Transport}

\subsubsection{Mean-Field Limit}

As the number of tokens $n \to \infty$, we describe the system 
using the empirical measure 
$\mu_s = \frac{1}{n} \sum_{i=1}^n \delta_{x_i(s)}$. The 
evolution of this density is governed by the continuity equation:
\begin{equation}
    \partial_s \mu_s + \nabla \cdot 
    (\mu_s\, \mathcal{X}[\mu_s]) = 0,
\end{equation}
where the vector field is
\begin{equation}
    \mathcal{X}[\mu](x) = \mathbf{P}_{x}^{\perp} \!\left( 
    \int e^{\beta \langle x, y \rangle} y \, d\mu(y) \right)
\end{equation}
for the unnormalized case.

\subsubsection{Maximizing Interaction Energy}

Both SA and USA dynamics are driven by the objective to 
\textbf{maximize the interaction 
energy}~\citep{rigollet2025mean,geshkovski2025mathematical}:
\begin{equation}
    \mathcal{E}_{\beta}(\mu) = \frac{1}{2\beta} 
    \iint e^{\beta \langle x, y \rangle} 
    d\mu(x)\, d\mu(y).
\end{equation}
While the objective is the same, the cost of movement differs 
between the two models:
\begin{enumerate}
    \item \textbf{USA as Wasserstein gradient flow:} The 
    unnormalized dynamics correspond to a gradient flow with 
    respect to the standard Wasserstein metric $W_2$. Attraction 
    can accelerate exponentially 
    ($v \propto e^\beta$)~\citep{rigollet2025mean,
    geshkovski2025mathematical}.
    \item \textbf{SA as weighted gradient flow:} Standard 
    self-attention corresponds to a gradient flow with respect to 
    a \textbf{weighted Wasserstein metric}. The normalization term 
    $Z$ cancels exponential growth, keeping the velocity field 
    bounded and regulating clustering 
    speed~\citep{rigollet2025mean,geshkovski2025mathematical}.
\end{enumerate}

\subsection{The Noisy Regime}
\label{app:input_perturbation}

\subsubsection{Continuous Noise (SDE)}

When noise is injected at every time step, the dynamics of a 
token $X_i(s)$ become a stochastic differential equation (SDE) 
on the sphere~\citep{rigollet2025mean}:
\begin{equation}
    dx_i(s) = 
    \underbrace{\mathbf{P}_{x_i(s)}^{\perp} \!\left( 
    \frac{1}{n} \sum_{j=1}^n 
    e^{\beta \langle x_i(s), x_j(s) \rangle} X_j(s) 
    \right) ds}_{\text{drift (clustering force)}} 
    + \underbrace{\sqrt{2\kappa^{-1}}\, 
    d\mathcal{W}_i(s)}_{\text{diffusion (noise)}},
\end{equation}
where $\kappa > 0$ controls the noise magnitude and 
$\mathcal{W}_i(s)$ denotes Brownian motion on the sphere, 
pushing particles apart.

\paragraph{Fokker-Planck dynamics and free energy.}
In the mean-field limit, the probability distribution $\mu_s$ 
evolves according to the Fokker-Planck equation, combining 
advective and diffusive 
transport~\citep{shalova2026solutions,rigollet2025mean}:
\begin{equation}
    \partial_s \mu_s = \nabla \cdot 
    \left( \mu_s \nabla \mathcal{V}[\mu_s] \right) 
    + \kappa^{-1} \Delta \mu_s,
\end{equation}
where $\mathcal{V}[\mu_s]$ is the potential field. The driving 
energy changes from pure interaction energy to \textbf{free 
energy}:
\[
    \text{Free energy} = \text{Interaction energy} 
    + \text{Entropy term}.
\]
\begin{itemize}
    \item \textbf{Low noise (high $\kappa$):} The clustering 
    force dominates; tokens group semantically.
    \item \textbf{High noise (low $\kappa$):} The diffusive 
    force dominates; the distribution becomes uniform and model 
    output can become incoherent.
\end{itemize}

\subsubsection{Input Perturbation}

Alternatively, noise can be added solely to the input as an 
\textbf{initial condition}, leaving the dynamics deterministic 
(ODE) for $s > 0$:
\begin{itemize}
    \item The initial condition is random: 
    $x_0 \sim \mu_0$.
    \item Due to the non-Lipschitz nature of attention and the 
    complex energy landscape (metastability), different 
    realisations of the initial noise can lead the system into 
    different basins of attraction.
    \item \textbf{Result:} Potentially different final 
    clusterings or attention patterns for the same semantic 
    input.
\end{itemize}

\end{document}